%% file: main.tex
\documentclass{article}
\usepackage{iclr2027_conference,times}
\input{math_commands.tex}

\usepackage[utf8]{inputenc}
\usepackage[T1]{fontenc}
\usepackage{amsmath,amssymb}
\usepackage{booktabs}
\usepackage{array}
\usepackage{graphicx}
\usepackage{xcolor}
\usepackage{hyperref}
\usepackage{url}
\usepackage{microtype}
\usepackage{multirow}
\usepackage{enumitem}
\usepackage{wrapfig}

\hypersetup{colorlinks=true,allcolors=black!70!blue}
\graphicspath{{./}}

\newcommand{\Delt}{\Delta}
\newcommand{\Enf}{E}
\newcommand{\Emyo}{E_{\mathrm{myopic}}}
\newcommand{\Enet}{E_{\mathrm{net}}}
\newcommand{\CD}{C_{\Delta}}
\newcommand{\CE}{C_{E}}
\newcommand{\fixed}{\textsf{fixed}}
\newcommand{\shuffled}{\textsf{shuffled}}
\newcommand{\high}{\textsf{visible}}
\newcommand{\low}{\textsf{hidden}}
\newcommand{\none}{\textsf{none}}
\newcommand{\freetext}{\textsf{free\_text}}
\newcommand{\BRone}{\mathrm{BR}_1}
\newcommand{\BRK}{\mathrm{BR}_K}
\newcommand{\ci}[2]{[#1, #2]}

\title{Persistent Partners Raise Prices \\ Among Learning Agents}

\author{Paul-Peter Arslan \\
Institute For Future Technologies \\
Devinci Higher Education \\
\texttt{paulpeterarslan@gmail.com} \\
\AND
Yubin Kim \\
Massachusetts Institute of Technology \\
\AND
Xiao Xiao \\
Institute For Future Technologies \\
Devinci Higher Education \\
}

\iclrfinalcopy 

\begin{document}
\maketitle
\lhead{Preprint. Under review.}

\begin{abstract}
When pricing agents meet repeatedly on a platform, the platform decides who faces whom. We ask whether that choice
moves the prices the agents learn, and whether a rise comes with learned punishment. In a pre-registered randomised
experiment in the Bertrand duopoly of Calvano et al., each agent's price is set by a tabular Q-learning module, not
by the small language model attached to it, and we randomise whether each agent keeps its partner, sees its rival's
prices and can send messages. Keeping the same partner raises the level of profits, averaged over training, by
$0.27$ of the gap between competitive and monopoly profit (95\% CI $0.20$ to $0.35$, all twenty paired runs
positive), our registered primary result, and the resting price by $0.17$ of the Nash-to-monopoly range (post hoc).
A plain tabular learner reproduces the effect in all 25 further blocks, and there one permanent partner raises the
level more than about three do ($+0.23$ against $+0.05$, exploratory). Where rival prices are hidden, the
price-setting module cannot see a cut, so cannot punish it, yet the resting price rises as much and the rise lasts to the end of training,
while with visible rivals it shrinks with longer training (post hoc). Where the rival is visible, a static best
responder accounts for a third to a half of what a forced-deviation probe reads as punishment, on the starts where
the rival can see the cut, and net of it the registered test of learned punishment is inconclusive. A test that
looks only for punishment would thus miss the rise where the rival is hidden, while a check for profitable
deviations flags most of those prices (post hoc). In an exploratory extension, untrained Qwen2.5 7B and 14B
models under one prompt show the effect when the rival's price is left out of the prompt and inconsistently when it
is shown, the 7B result replicating on fresh blocks, while two other model families show none.
\end{abstract}

\section{Introduction}
\label{sec:intro}

Algorithmic pricing is a live policy question \citep{harrington2018developing,assad2024algorithmic}, and collusion
is one of the named risks of interacting AI agents \citep{hammond2025multiagent}. Firms can now
delegate pricing to learning agents, including agents built on large language models (LLMs), and such agents meet the same
competitors again and again on a platform. When their prices come out too high, a regulator needs to know what in
their environment put them there, and whether the usual evidence of coordination, a rival that punishes a price cut,
would be there to find. We study one property of that environment that neither firm controls, whether an agent keeps
meeting one partner or is spread over several. Our central result is that keeping the same partner raises the prices agents learn. We
establish it in a randomised, pre-registered experiment on hybrid agents whose price is set by a tabular Q-learning
module, not by the attached language model, and we find it again, in a separate exploratory extension, in untrained
prompted Qwen2.5 models, with rematched ones re-paired every round, when the rival's price is left out of the prompt. The usual explanation for high prices between firms that meet
repeatedly is that each can punish a deviation. In half of our conditions the rival's prices are hidden, and where
there is also no message channel, no trained policy can condition its price on a deviation. Keeping the same partner still raises prices there, in a
breakdown we did not register. There punishment is not merely undetected but impossible, so a test looking only for
punishment would find nothing, while a check for profitable deviations flags most of those rest points (post hoc), as the regret audit of
\citet{hartline2024regulation} is designed to.

\citet{fish2024algorithmic} showed that \emph{prompted} language
models reach supra-competitive prices in a calibrated Bertrand duopoly, and that each agent's price moves with its
rival's previous price in a way they read as reward-punishment. Their models are not trained, so their design cannot say what a training process contributes, and work on
algorithmic collusion has mostly varied the algorithm, which the firms choose (Section~\ref{sec:related}). Studies
of Q-learning pricing let the same learners meet throughout, so what we add to them is a randomised test of who
faces whom, an enforcement endpoint net of the static response, and higher prices where no policy can retaliate.
We ask the causal question instead. \emph{Which properties of the environment in which agents learn against each other
move the prices they reach, and do the same properties move enforcement?}

\paragraph{Threat model.} Two rival firms each delegate pricing to a learning agent, and the two agents transact
repeatedly in one marketplace. Nobody instructs either agent to collude, neither
firm is an adversary, and each agent maximises its own principal's profit. No message passes between the agents
unless a channel is provided. The harm is a price above the competitive level, and it is paid by consumers. Whether
an agent keeps trading with one partner or is rematched among several is decided by the marketplace operator, not
the firms, and platform rules
can be designed against algorithmic collusion \citep{johnson2023platform}. That lever exists where a platform
assigns counterparties and not where geography fixes the rival.

The two endpoints are the level~$\Delt$ of profits and the rival's response to a forced deviation \emph{net of its
static best response}, $\Enet$, the correction being needed because a static best-responder also cuts
its price when its rival undercuts it. The design randomises three properties of the environment. An agent either
keeps its partner or is rematched at every episode end, the two \emph{arms} \fixed{} and \shuffled{}, the rival's
last prices are \high{} or \low{}, and a free-text channel exists or does not (\none{} / \freetext{}). Each
combination is a \emph{cell}. We analyse forty runs in five seed
blocks by randomisation inference, with the hypotheses, statistic and error budget registered with a frozen copy of
the code before the first confirmatory run \citep{nosek2018preregistration}.

\paragraph{What we found (registered).}
The primary registered test rejects, with all twenty paired runs positive. Keeping the same partner raises the
level, averaged over training, by $+0.27$, $\ci{+0.20}{+0.35}$, on a profit scale where $0$ is what two firms earn
when neither restrains itself (Nash) and $1$ what a single owner of both would earn, with random play at $0.38$, and
the resting price by $0.17$ of the Nash-to-monopoly price range (post hoc, also positive in all twenty run pairs).
The whole interval lies above the smallest effect we registered as worth detecting, and a plain tabular learner in
the same design reproduces the effect in all 25 further blocks (exploratory). There one permanent partner per
learner raises the level by $+0.23$, $\ci{+0.13}{+0.33}$, and about three permanent partners by $+0.05$,
$\ci{-0.07}{+0.17}$, so what matters seems to be meeting one partner rather than how long a pairing lasts, though
that interval does not show that three partners equal full rematching (exploratory). In the post hoc split, the rise lasts
to the end of training where the rival's price is hidden ($+0.52$ at the last checkpoint), and where it is visible
it is not distinguishable from zero there. Where the rival is visible, a static best responder accounts for a third
to a half of what an uncorrected probe reads as punishment on the starts where the rival can see the cut, and more
than all of it in one cell, and net of it the
registered test on enforcement
is inconclusive, planned at a power of about one half (Section~\ref{sec:confirmatory}). A second registered test,
pairing trained agents with an agent they did not train with, also rejects, but it cannot separate adaptation to one
partner from the general cost of meeting someone new.

\paragraph{Which part of the agent responds, and an extension (exploratory).}
Experiments with predictions committed before their runs suggest that the value module carries the level effect,
since zeroing the language model's head leaves the contrast unchanged in the one cell tested. When the head overturns
the table at 49\% and 79\% of \emph{settled seats}, where the probed pair settles on one price pair, both arms fall
and the contrast survives, mostly because rematched agents price lower, and the backbone trained alone by policy
gradient shows no clear effect. As a separate extension, untrained Qwen2.5 7B and 14B models prompted with their own history, rematched ones
re-paired every round, show the effect in all sixteen blocks when the rival's price is left out of the prompt and
inconsistently when it is shown. The 7B result replicates on fresh blocks and keeps about half its size when each
agent sees only its older rounds with its current partner, while Mistral-7B and OLMo-2-7B price high in both arms
and show none (Section~\ref{sec:parts}).

After seeing four of the five blocks we chose a candidate explanation, \emph{spontaneous coupling}
\citep{banchio2022spontaneous}, and tested it with predictions committed before their own runs or analyses. Two
learners that keep meeting learn from the same run of prices, so when one tries a high price and is paid well, the
other is paid well at the same moment for whatever it was doing, and their value estimates drift upward together,
with neither reacting to the other or holding a threat. About half of those predictions held by direction and one
by intervals at five blocks (Section~\ref{sec:mechanism}).

\paragraph{Contributions.}
\begin{enumerate}[leftmargin=1.4em,itemsep=1pt,topsep=2pt]
\item \textbf{A pre-registered randomised test of a lever that platforms hold}, who faces whom, which studies of
  Q-learning pricing keep fixed. Keeping the same partner raises the prices these agents reach, in all twenty paired
  runs and in all 25 replication blocks of a plain tabular learner, where one permanent partner raises them more
  than about three (exploratory, Sections~\ref{sec:setup} to \ref{sec:mechanism}).
\item \textbf{Higher prices where retaliation is impossible}. Where the rival's prices are hidden the module that
  sets the price cannot see a cut, yet the resting price rises as much as overall and the rise lasts to the end of
  training (post hoc, Section~\ref{sec:confirmatory}).
\item \textbf{An enforcement endpoint} net of the static best response, with a proof that a rival following the
  matched static reference with inertia, delay or smoothing scores zero in the limit, and evidence that uncorrected
  probes read ordinary best responses as punishment, a third to a half of the raw response on the starts where the
  rival can see the cut and half or more over all starts. Its registered test was planned at a power of about one
  half, and correcting the probe's blind starts raises its estimate (post hoc, Sections~\ref{sec:endpoints}
  and \ref{sec:confirmatory}).
\item \textbf{A mechanism tested by predictions committed in advance}, consistent with, but not establishing, the
  coupling of two learners' value estimates (Section~\ref{sec:mechanism}).
\item \textbf{An exploratory extension to prompted language models}, with predictions committed before each run.
  Untrained Qwen2.5 models show the effect with the rival's price hidden, the 7B result replicated on fresh blocks,
  and two other families show none (Section~\ref{sec:parts}).
\end{enumerate}

\section{Related work}
\label{sec:related}

\paragraph{Q-learning and the origin of the environment.}
\citet{calvano2020ai} established that tabular Q-learners reach supra-competitive prices in this calibrated
duopoly and read reward-punishment off a forced deviation. Our table uses their rule on a coarser state with
faster-decaying exploration. Later work mapped what governs the phenomenon among Q-learners
\citep{klein2021autonomous,xu2024memoryless,ye2025demand,asker2022artificial,asker2024impact,calvano2021imperfect,abada2023artificial},
and faster, reactive pricing algorithms raise prices even in a competitive equilibrium \citep{brown2023competition}.
Auction formats matter for Q-learning bidders \citep{banchio2022auction}. No study of algorithmic pricing we know of randomises who faces whom, and that
contrast, with an endpoint that subtracts the static response, is what is new here.

\paragraph{Partner persistence in multi-agent learning.}
Failing against a fresh partner after training with a fixed one is a known failure mode in multi-agent
reinforcement learning, treated as co-adaptation, measured by cross-play \citep{hu2020otherplay,lupu2021trajectory}
and addressed by training against a diverse pool of partners \citep{strouse2021collaborating}, since agents co-trained
through random pairings still overfit to their training partners. A varied population changes the policies
agents reach in social dilemmas \citep{mckee2020social}, and steering a partner's learning toward cooperation
\citep{foerster2018lola} presumes that the partner stays. Our \shuffled{} arm is such random pairing and our swap-opponent
probe is that evaluation. Here keeping the partner raises the price, though our tests do not show that adaptation
to that partner is the cause.

\paragraph{Supra-competitive prices without threats.}
Supra-competitive prices arise without any threat from Q-learners in Cournot markets
\citep{waltman2008qlearning} and, as above-Nash payoffs from some starting values, in stateless repeated dilemmas
\citep{wunder2010classes}, from independent algorithms that ignore rivals \citep{hansen2021algorithmic}, only transiently for least-squares
learners \citep{wu2026oblivious}, from a
no-regret first mover \citep{arunachaleswaran2024threats}, and from \emph{spontaneous coupling}
\citep{banchio2022spontaneous}, which disappears under frequent experimentation and does not occur in the
synchronous learner of \citet{asker2022artificial}. \citet{calvano2023genuine} separate such prices from genuine
reward-punishment. The coupling account implies that learners must keep meeting, and our contrast tests that
implication rather than a new prediction. Without observed rival actions, collusion needs punishments triggered by
market outcomes \citep{green1984noncooperative}, yet our effect lasts where the agent's state holds no such signal. Audit proposals test an algorithm for
reward-punishment \citep{harrington2018developing} or for calibrated regret \citep{hartline2024regulation}.

\paragraph{LLM agents.}
Beyond \citet{fish2024algorithmic}, others optimise a \emph{shared} prompt, vary data access and patience at test
time, or randomise interventions on frozen agents \citep{tian2026prompt,keppo2026fragility,riedl2026emergent}.
Others train LLM agents in social dilemmas and strategic conversation
\citep{piche2025robust,conchello2026gametalk}, as deep Q-learners were earlier trained in sequential social dilemmas
\citep{leibo2017multi}, varying the algorithm and the game.
Work on covert channels measures collusion as a capability \citep{motwani2024secret} and as a propensity
\citep{nakamura2026colosseum} or finds
it emerging from misspecified training incentives \citep{mathew2024hidden}. \citet{weis2026incontext} obtain
cooperation from in-context inference of the co-player, which on our reading could substitute for a persistent
partner and predicts a \emph{smaller} persistence effect, a rival prediction we registered. Cheap talk fails among
purely self-interested learners \citep{cao2018negotiation,lazaridou2020emergent}. Random
rematching is a classic treatment in human experiments \citep{andreoni2008partners,duffy2009cooperative,dalbo2018determinants},
where players observe what the others in their match did, so any extra cooperation among partners can be read as enforcement, and in
theory cooperation under random rematching rests on community enforcement
\citep{kandori1992social,ellison1994cooperation}, which a hidden rival rules out.

\section{Experimental setup}
\label{sec:setup}

\subsection{Environment}
We use the calibrated differentiated-product Bertrand duopoly of \citet{calvano2020ai}, namely logit demand with
differentiation $\mu = 0.25$, two firms and simultaneous moves. The grid of 33 prices, with the static Nash equilibrium at bin~6 and the
joint-monopoly price at bin~26, and the episodes ending with continuation probability $0.98$ are our choices (their baseline has 15 prices and no episodes). Sixteen
two-seat markets run in parallel per run, and a run lasts 8\,000 rounds with a checkpoint every 1\,000. All eight
cells start
from the same table, Calvano's initialisation, so the initial greedy price is the static best response to a uniform
rival in every cell (bin 11, $\Delt = 0.39$, Appendix~\ref{app:agent}).

\subsection{The agent, a language backbone coupled to a value module}
\label{sec:agent}
The agent has two parts. The first is a \textbf{language backbone}, Qwen2.5-0.5B-Instruct \citep{qwen2025qwen25},
with LoRA adapters \citep{hu2022lora} and a linear price head on the centred hidden state. We train it by policy
gradient \citep{williams1992simple}. The second is a
\textbf{tabular value module}, a $4\times4$ table over (bucketed last rival price, bucketed last own price), plus one
no-history state and four \emph{rival-unobserved} states used when the rival is hidden. The table is learned off-policy by the
Calvano rule $Q(s,a) \leftarrow (1-\eta_Q)Q(s,a) + \eta_Q\,(r + \delta\max_{a'}Q(s',a'))$ with $\eta_Q = 0.15$ and
$\delta = 0.95$. Action logits are the
head's output plus $16.5\,[\mathrm{onehot}(\arg\max_a Q(s,a)) + 0.5\,\widetilde{Q}(s,\cdot)]$, with $\widetilde{Q}$
the min-max normalised row, so the table's best action has a lead of at least $16.5$ logits. Behaviour is
$\varepsilon$-uniform with $\varepsilon = 0.3\times0.9994^{t}$. Hyper-parameters are in Appendix~\ref{app:agent}.

The two-part design follows a negative result. Trained alone on-policy by policy gradient, the
backbone reached no collusive point (Section~\ref{sec:parts}), nor did a tabular toy with its learning rule,
even when handed the relevant state, while the Calvano rule reaches such points on the same table, so at this
budget the obstacle is the learning rule, not the environment (Appendix~\ref{app:phase1}). Adding the value module
is what let the hybrid reach such points, and the table's weight, $16.5$, sets how far it leads the head
(Section~\ref{sec:parts}). With the head's logits capped at
$c = 5.5$ it can separate two actions by at most 11 logits against the table's lead of at least $16.5$, so it never
changes the greedy price that the probe reads, and during training it moves only a small share of the sampling
probability.

\subsection{Three factors, defined as availability}
\label{sec:factors}
\textbf{Partner persistence.} Under \fixed{}, agents A and B trade with each other in eight markets and C and D in
the other eight, for the whole run. Under \shuffled{}, all markets end their episodes at the same time, and the
seats are re-paired at random at every episode end. Each market always holds two distinct agents, and the price
history belongs to the \emph{seat} and stays with it. \textbf{Observability.} When the rival is \high{}, its last prices
appear in the prompt and in the table's state. When it is \low{} they appear in neither. A greedy policy cannot react to a deviation it cannot see, so there every enforcement
quantity is zero by design and not by finding, and what we test there is the level alone.
\textbf{Communication.} Under \freetext{}, each agent emits a 6-token message per round,
trained by policy gradient with a KL penalty to the adapter-free backbone.
\shuffled{} needs at least four agents ($M = 4$ here) to differ from \fixed{}, since with two a rematched learner meets the same
partner again.

In a check on one block, the trained sender and untrained senders of up to 3B parameters wrote the same sentence
whatever the market showed, so the channel gives the price nothing to use
(Appendix~\ref{app:explo}). Under \shuffled{}, policy-gradient updates are also $5.9$ times fewer, and larger, on
the same number of transitions, and an agent's experience is spread over three partners, both counted in the
treatment (Section~\ref{sec:limitations}).

\subsection{Blocks, slot seeds and the random assignment}
Eight cells $\times$ five seed blocks give 40 runs. The eight cells of a block share a block seed and so start
from identical parameters, and a block RNG assigns eight \emph{slot seeds}, which control sampling, exploration,
episode ends and re-pairing, to cells by a permutation printed into the launch script before the block runs. The
analysis inverts that assignment (manipulation checks in Appendix~\ref{app:explo}).

\section{Endpoints and inference}
\label{sec:endpoints}

Table~\ref{tab:notation} collects the symbols.

\begin{wrapfigure}{r}{0.5\linewidth}
\vspace{-12pt}
\centering
\includegraphics[width=\linewidth]{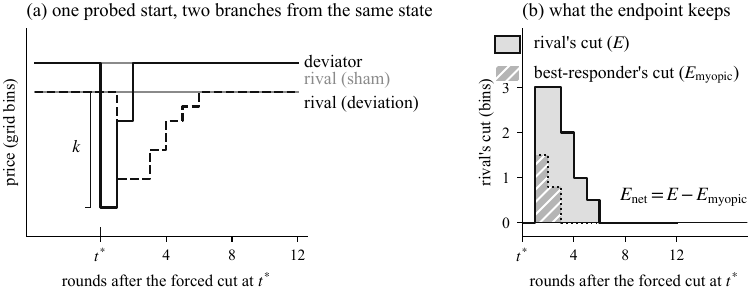}
\caption{The probe and the correction. (a) The sham and deviation branches from a rest point. (b) The rival's cut has area $\Enf$. A static best-responder
on the same deviator path makes its own cut ($\Emyo$, hatched), and $\Enet$ is the difference.}
\label{fig:probe}
\vspace{-10pt}
\end{wrapfigure}
\paragraph{The forced-deviation probe.}
At every checkpoint we freeze the policies and probe them, following
\citet{calvano2020ai} (Figure~\ref{fig:probe}a). From each of four starting
histories drawn on the grid, the frozen pair plays 40 greedy rounds and settles at what we call a \emph{rest point}.
The run then splits from that state into a control branch with no intervention, the \emph{sham} branch, and a
\emph{deviation} branch in which one firm is forced to $k = 4$ bins below its rival's price, and we watch 25 rounds. Prices are read
greedily, so the probe measures the greedy policy and not the policy as played. The probed unit is a
\emph{point}, one probed pair at one checkpoint, and with two pairs per checkpoint there are 16 points per run. Its
level $\Delt = (\pi - \pi_{\mathrm{Nash}})/(\pi_{\mathrm{mon}} - \pi_{\mathrm{Nash}})$ is taken on the \emph{profits}
at the rest point, averaged over firms and starts.

\paragraph{Enforcement net of the static response.}
The raw score is the signed area between the branches (Figure~\ref{fig:probe}b),
$\Enf = -\mathrm{AUC}_{h=1..25}\,\mathrm{RE}(h)$, with $\mathrm{RE}$ the rival's deviation-branch price minus its
sham-branch price in bins. Positive $\Enf$ means the rival went lower after the deviation than it otherwise
would have. For $r \in \{\BRone, \BRK\}$ ($K = 5$ rounds of history), $\Emyo^r$ is the same area, computed for a
static best-responder that we apply round by round to the deviator's \emph{realised} price path in both branches,
and which never enters the market. The endpoint is $\Enet^r = \Enf - \Emyo^r$, in bins$\cdot$rounds.

\paragraph{Properties of the correction.}
The benchmark rules out one error only, since $\Enet > 0$ does not by itself establish strategy. By a theorem
proved in Appendix~\ref{app:validity}, if the rival's price is the matched static reference passed through a
causal, absolutely summable linear filter with static gain $G$, with a shared pre-history, then
$\Enet = (G-1)\,\Emyo$, so a rival that merely follows the static best response with inertia, delay or smoothing,
all of unit gain, scores zero in the limit, and about zero over our 25 rounds. A best response to a \emph{window} of past prices is not covered. $\BRK$ handles it, but the two benchmarks
leave biases of opposite signs, so H6 tests both. Two further biases, which neither benchmark removes, push $\Enet$
down. A rival that punishes keeps the deviator low, which also enlarges $\Emyo$, and on the $43\%$ of starts whose
cut leaves the deviator's price bucket unchanged the rival cannot respond, so $\Enet = -\Emyo$ there
(Appendix~\ref{app:explo}).

\paragraph{The descriptive verdict.}
We call a point \emph{collusive}, the name we registered, only when three conditions hold together. An uncorrected punishment pattern, a cut of at least
one bin followed by at least 50\% recovery within 25 rounds as in the punishment phase of
\citet{abreu1988theory}, appears on at least half the starts. The level exceeds $0.6$. And the spread of $\Delt$
between the two firms is below $0.3$. No error budget is spent on this verdict.

\paragraph{Hypotheses and error budget.}
Run-level outcomes are means over the 16 points, with no checkpoint selected. Let $d_{b,O,C}$ be the difference
between the \fixed{} and \shuffled{} runs in block $b$ at observability $O$ and channel $C$. \textbf{H1} (primary)
states that $\CD = \tfrac14\sum_{O,C} d_{b,O,C}$, averaged over blocks, is positive, and is tested one-sided at
$\alpha_\Delt = 0.030$. \textbf{H6} (secondary) takes the same contrast on
$\Enet^{r}$, rejected only if \emph{both} benchmarks reject at $\alpha_E = 0.010$. \textbf{Stage 2} adds two
confirmatory probes on the frozen final policies, which share $\alpha_M = 0.010$ under Holm's procedure so that the
smaller $p$-value must pass $0.005$, \emph{mask rival history}, which hides the rival's prices from both agents, on
$\Enet$, and \emph{swap opponent} (pairs (A,C), (B,D) instead of (A,B), (C,D)) on $\Delt$, both predicted to
decrease. \emph{Mute messages}, the interaction
$O\times C$ (H2) and a channel contrast (H3) are registered as exploratory (Appendix~\ref{app:explo}).

\paragraph{Inference.}
The statistic is $T = \mathrm{mean}_b(C_b)/(\mathrm{sd}_b(C_b)/\sqrt{5})$ with $C_b$ the block contrast, and the
reference distribution flips the sign of each $d_{b,O,C}$ independently, the $2^{20}$ patterns enumerated exactly.
The test is exact under the sharp null of no effect in any run pair, and studentising the statistic is meant to keep it
asymptotically valid for the weak null, as it does for two-sample permutation tests \citep{chung2013exact}, only
approximately at five blocks. Confidence sets
invert the two-sided test on a grid
(Appendix~\ref{app:inference}). Each result then falls into one of four verdicts fixed in advance,
\emph{substantial} (rejection and interval beyond the smallest effect size of interest, or SESOI), \emph{precise
null} (no rejection, interval within $\pm$SESOI), \emph{reversed} (interval below $-$SESOI) or
\emph{inconclusive} (interval crossing a boundary).

\section{Confirmatory results}
\label{sec:confirmatory}

\begin{figure}[t]
\centering
\includegraphics[width=\linewidth]{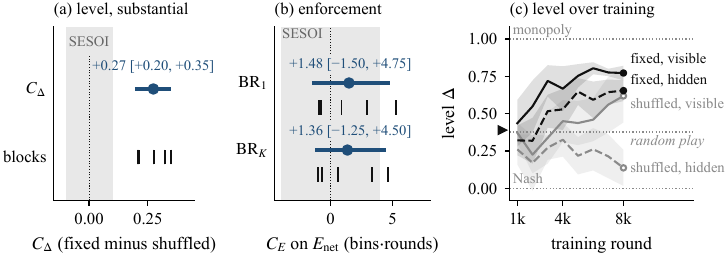}
\caption{The two registered contrasts and the level profile, five blocks. (a, b) Run-level means over the eight
checkpoints. Dot and bar give the point estimate and the 95\% confidence set by test inversion, ticks the five block
contrasts, and the grey region $\pm$SESOI. (c) Resting level by checkpoint, mean and 95\% band of the 10 runs per
line, and a marker at the common start of $0.39$ (exploratory).}
\label{fig:main}
\end{figure}

The registered analysis ran on the five complete blocks, and again after a bug fix involving no analytic choice let
the swap-opponent test run, with H1 and H6 bit-identical (Appendix~\ref{app:deviations}).

\begin{table}[h]
\centering\footnotesize
\caption{Results at a glance. Registered tests first ($p$ one-sided), then post hoc readings that spend no $\alpha$. Stage 2 rows instead compare probed with standard final
policies. Intervals are two-sided 95\% sets. Levels are run-level means over the eight checkpoints unless marked
final.}
\label{tab:results}
\setlength{\tabcolsep}{3pt}
\begin{tabular}{@{}llrrrl@{}}
\toprule
contrast (\fixed{} minus \shuffled{}) & status & estimate & 95\% CI & $p$ & verdict \\
\midrule
H1, level $\CD$ & registered & $+0.274$ & $\ci{+0.195}{+0.350}$ & $0.00034$ & substantial \\
H6, enforcement $\CE^{\BRone}$ & registered & $+1.48$ & $\ci{-1.50}{+4.75}$ & $0.145$ & inconclusive \\
H6, enforcement $\CE^{\BRK}$ & registered & $+1.36$ & $\ci{-1.25}{+4.50}$ & $0.156$ & inconclusive \\
swap, \fixed{} agents vs a non-partner, $\Delt$ & stage 2, Holm & $-0.177$ & $\ci{-0.275}{-0.080}$ & $0.0003$ & rejected \\
mask rival history, $\Enet$ & stage 2, Holm & $-0.70$ & $\ci{-10.75}{+10.75}$ & $0.43$ & not rejected \\
\midrule
level, rival hidden & post hoc & $+0.295$ & $\ci{+0.240}{+0.350}$ & -- & -- \\
level, final checkpoint & post hoc & $+0.335$ & $\ci{+0.185}{+0.485}$ & -- & -- \\
\quad rival hidden, final & post hoc & $+0.52$ & $\ci{+0.42}{+0.62}$ & -- & -- \\
\quad rival visible, final & post hoc & $+0.15$ & $\ci{-0.13}{+0.46}$ & -- & -- \\
hidden minus visible & post hoc & $+0.04$ & $\ci{-0.03}{+0.14}$ & -- & -- \\
hidden minus visible, final & post hoc & $+0.36$ & $\ci{+0.18}{+0.56}$ & -- & -- \\
\begin{tabular}[t]{@{}l@{}}resting price, share of the\\ Nash-to-monopoly price range\end{tabular} & post hoc & $+0.17$ & $\ci{+0.12}{+0.22}$ & -- & -- \\
\bottomrule
\end{tabular}
\end{table}

\paragraph{H1, the level, rejected and substantial.}
The interval for $\CD$ lies entirely above $\mathrm{SESOI}_\Delt = 0.10$ ($T = 9.56$), and all five block contrasts
(Figure~\ref{fig:main}a) and all twenty run pairs are positive. At the final checkpoint the visible-rival
contrast is not distinguishable from zero (Table~\ref{tab:results}, Figure~\ref{fig:main}c), so with a visible
rival much of the registered effect reflects how fast prices rise (post hoc).

\paragraph{H6, enforcement, not rejected and inconclusive.}
Neither component rejects at $\alpha_E = 0.010$, and both intervals cross $\mathrm{SESOI}_E = 4$ at their upper end,
so the registered verdict is \emph{inconclusive}. The data neither show that persistence raises
enforcement nor bound its effect below what would matter. Removing the bias of the blind starts raises
$\CE^{\BRone}$ from $+1.48$ to $+1.72$, positive in three blocks of five as before (post hoc, no interval,
Appendix~\ref{app:validity}).
Power at the smallest effect of interest was about one half as planned, and at most
$0.48$ recomputed from the observed spread (Appendix~\ref{app:inference}). As enforcement is zero by construction where the rival
is hidden, the registered four-cell average is exactly half the visible-rival contrast, which is $+2.97$ and $+2.71$
on a scale where the smallest effect of interest is $8$.

\paragraph{What the correction does to the reading.}
In all four visible-rival cells the net response is smaller than the raw one (Table~\ref{tab:percell}). A static
best-responder accounts for 48\% of the raw response in \fixed{}/\high{}/\none{}, the cell where it is largest
($\Enf = 6.99$ bins$\cdot$rounds against a net $3.61$), for 71\% and 51\% in two others, or 40\%, 50\% and 32\% on
the starts where the rival can see the cut, and in \shuffled{}/\high{}/\freetext{} the sign flips, from $5.16$ to
$-2.04$.

\paragraph{Level and enforcement come apart where the rival is hidden.}
These results support a narrower claim than ``collusion''. Of the 640 probed points,
54 (8.4\%) are collusive by the descriptive verdict, with its uncorrected punishment pattern, all with a visible
rival, as the pattern cannot occur on the 320 hidden-rival points. Yet persistence raises
the level and the resting price there as well (level $0.53$ against $0.23$, resting price $1.66$ against $1.58$,
with Nash at $1.47$ and monopoly at $1.92$, a rise of $0.17$ of that range, $\ci{+0.12}{+0.23}$, post hoc), with all ten
block-by-channel differences positive, one arm rising and the other falling from their common start of $0.39$. Most rest points there are
not equilibria, since at 80\% of starts for persistent pairs and 60\% for rematched ones a four-bin cut pays at once
and goes unanswered (post hoc). The hidden-rival contrast exceeds the visible-rival one at the final checkpoint but
not averaged over checkpoints (Table~\ref{tab:results}, post hoc).

\paragraph{Stage 2, the swap-opponent test.}
\emph{Swap opponent}, which pairs frozen final \fixed{} agents with a non-partner, is \textbf{rejected} ($p = 0.0003$, Holm threshold $0.005$, $n = 20$). By block instead of by run (post hoc), all five block means are negative
(one-sided $p = 1/32$, the smallest five blocks allow). The drop, $0.18$ on average, is $0.33$ where the
rival is visible, most of the final level's excess over random play, and $0.02$ where it is hidden. Within a run,
the same
pairs of \shuffled{} agents met during training, so the near-zero drop there ($-0.01$ on the hybrid) is not a test
against strangers. Against true strangers from another block, on the tabular learner, rematched agents lose $0.18$
and persistent ones $0.28$, a difference of $-0.10$, $\ci{-0.26}{+0.07}$, so our later prediction that persistent
agents lose more held by direction and failed by its interval rule (Appendix~\ref{app:tabrev}). The test therefore
cannot separate partner-specific adaptation from the cost of facing a stranger, and on the tabular learner, the only
agent we tested against true strangers, the point estimates attribute most of the drop to the latter. \emph{Mask
rival history}, confirmatory, does not reject on either benchmark ($n = 10$).

\section{Mechanism, partly consistent with spontaneous coupling}
\label{sec:mechanism}

A level effect without established enforcement, even where punishment cannot occur, is what spontaneous coupling
predicts (Section~\ref{sec:related}). Table~\ref{tab:scorecard} scores what we predicted would switch it off, all
exploratory.

\paragraph{The same design with a pure tabular learner.}
\label{sec:baseline}
A plain Calvano Q-learner, identical to the agent's table, in the same design and probe, reproduces a larger
persistence effect, $\CD = +0.363$, $\ci{+0.215}{+0.520}$, and $+0.306$,
$\ci{+0.270}{+0.340}$, on 25 further blocks (Appendix~\ref{app:tabrev}). Its enforcement contrast is $-2.58$,
$\ci{-7.25}{+1.75}$, on the five blocks ($-5.16$ on the visible-only scale), while on the 25,
where the rival is visible, the sign reverses, $+3.75$, $\ci{+0.20}{+7.30}$, against $\BRone$ and $+3.15$,
$\ci{-0.45}{+6.70}$, against $\BRK$, not established by the two-benchmark rule of H6. Both upper bounds lie below $8$, the smallest
effect of interest, and a deeper cut gives intervals above zero on both, so with a visible rival the value learner
may learn some enforcement, and the clean dissociation holds where the rival is hidden.
Run to $10^6$ rounds, this learner and the agent's own value
module end with a large gap where the rival is hidden ($+0.77$, $+0.59$, the first's rematched arm ending
at $-0.090$, Table~\ref{tab:mechcells}) and a small one where it is visible ($+0.22$, $+0.08$), where the
rematched learner shows the punishment pattern most often of any tabular cell (Appendix~\ref{app:mechanism}).

\paragraph{Switching the effect off.}
Coupling needs learners who keep meeting, so it should vanish when a learner has a single possible partner. With two
agents, rematching changes only the shared episode clock and the seat's history, both arms sit at the \fixed{}
level, and the contrast is $-0.032$, $\ci{-0.190}{+0.130}$, and $-0.025$, $\ci{-0.065}{+0.015}$, on 25 further blocks,
a precise null, so in the tabular learner the partner's identity, not these confounds, produces the effect. The
synchronous learner of \citet{asker2022artificial}, which by the coupling account cannot couple, shows a small negative effect ($-0.042$, and $-0.036$, $\ci{-0.065}{-0.010}$, on 25
blocks), though its levels do not land near Nash as predicted. Eight agents and a permanent
exploration floor move the contrast the predicted way ($+0.396$ and $+0.264$ against $+0.363$), with overlapping
intervals, and rematching a partner with probability $0.25$ per episode already gives the \shuffled{} level
(Figure~\ref{fig:mechanism}). Set side by side, one permanent partner per learner gives a contrast with full
rematching of $+0.229$, $\ci{+0.125}{+0.325}$, and about three permanent partners in an unplanned variant, $+0.048$, $\ci{-0.065}{+0.170}$ (Appendix~\ref{app:mechanism}), so facing one partner,
rather than a long pairing, seems to raise prices, though that interval does not show that three partners equal
full rematching (exploratory, five blocks).

\paragraph{The signature of coupling.}
\label{sec:signature}
Under coupling, learners that keep meeting should grow similar value tables. The hybrid agents' table correlation, partners minus
non-partners, grows under \fixed{} and stays at zero under \shuffled{}, strongly with a hidden rival ($+0.17$ without
a channel, $+0.05$ with one, at 8\,000 rounds) and slightly with a visible one ($+0.013$, where all tables correlate
at about $0.9$). A rival account, that a learner meeting several partners best-responds to their average, more
competitive policy, also fits the switch-off tests, the three-partner variant included, and the signature separates
the two only in part. On 25 tabular blocks, pairs that never met correlate at $+0.001$ and $+0.017$ against $+0.025$
and $+0.036$ for partners (rival visible, hidden), so where the rival is hidden about half of the signature is not
specific to partners, and our placebo prediction failed in both cells.

\section{Which agents respond to persistence}
\label{sec:parts}

\begin{figure}[t]
\centering
\includegraphics[width=\linewidth]{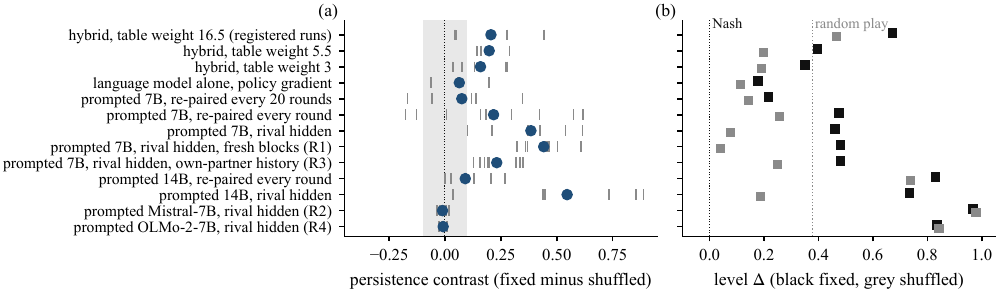}
\caption{Who responds to persistence, rival visible, no channel, unless marked. (a) Block contrasts (ticks),
mean (dot), band $\pm\mathrm{SESOI}_\Delt$. (b) Levels, \fixed{} black, \shuffled{} grey. Hybrid rows average eight
checkpoints, the language model four, prompted agents the last 100 rounds (50 if re-paired every 20), so
rows compare in sign and size. Planned blocks (7B replication, rival shown, $+0.01$).}
\label{fig:agents}
\end{figure}

Figure~\ref{fig:agents} and Table~\ref{tab:parts} vary who sets the price, predictions labelled by experiment
(Appendices~\ref{app:validity}, \ref{app:lmagents}).

\paragraph{The value module, and the language model given authority.}
With the head zeroed and frozen in the visible-rival cell without a channel, the value module alone sets the price
and the contrast is unchanged, $+0.204$ against $+0.206$, though the swap drop is smaller (D3 failed,
Appendix~\ref{app:validity}). A table weight of $5.5$, not $16.5$, lets the head overturn the table in 49\% of settled seats, against 0 of 408
in the ablation. On the three planned blocks both arms fall, to $0.395$ and $0.198$, a contrast of $+0.198$, and it
stays positive on two added blocks (F1, F2 held). At weight $3$, on five planned blocks, the head
overturns it at 79\% of settled seats and the contrast is still $+0.159$, 5 of 5, a smaller drop than the $0.05$ we
predicted (G2$'$ failed, Appendix~\ref{app:validity}). The \fixed{} arm ends below its start of $0.39$ at weight $3$
and near it at $5.5$, so where the head decides most prices, rematching lowers them rather than persistence raising them. Trained alone by policy gradient on three blocks, the
backbone prices near Nash ($0.18$, $0.11$), a contrast of $+0.06$ against $+0.26$ for the hybrid on those blocks and checkpoints
(E1 held, descriptive, maybe a floor).

\paragraph{An exploratory extension, prompted language models.}
Prompted with its own last 30 rounds over 500 rounds, the rematched arm re-paired every round among four agents, an
untrained model responds when the rival's price is left out of the prompt, though its own sales reveal that price
exactly, so these runs do not separate level from enforcement.
Over the last 100 rounds the 7B model gives $+0.38$, $\ci{+0.24}{+0.52}$, and the 14B model $+0.55$,
$\ci{+0.31}{+0.79}$, each positive in 8 of 8 planned blocks, the 7B against our prediction and the 14B as
predicted (L1). On eight fresh blocks the 7B model
gives $+0.44$, $\ci{+0.37}{+0.53}$, 8 of 8 (R1 held). Its persistent arm sits only
slightly above random play ($0.46$ to $0.48$ against $0.378$), so most of the contrast comes from rematched agents
pricing far lower. Re-pairing also mixes several rivals in an agent's history. Shown only its older rounds with
its current partner, the rematched arm rises from $0.04$ to $0.25$, so mixing or staleness accounts for about half
the gap, and the persistent arm still leads by $+0.23$, $\ci{+0.16}{+0.33}$, 8 of 8 (R3 held). Mistral-7B sits at
a monopoly ceiling ($0.97$, $0.98$) and OLMo-2-7B at $0.84$, a precise null ($-0.01$, $\ci{-0.02}{0.00}$), so R2
and R4 failed, and we did not test whether these two respond to their environment under this prompt. With the rival's price shown
the 7B evidence is mixed, $+0.22$, $\ci{+0.02}{+0.42}$, on ten planned blocks but $+0.01$, $\ci{-0.14}{+0.16}$,
on ten added ones. Earlier designs, adapted after each reading, are all reported in Appendix~\ref{app:lmagents},
uncorrected for multiple comparisons.

\section{Limitations and conclusion}
\label{sec:limitations}

Every contrast is conditional on one environment with $M = 4$ and 8\,000 training rounds, one trained 0.5B model and
prompted models under one prompt, of which one family of three shows the effect. With a visible rival the gap mostly
reflects learning speed, and hiding the rival also shrinks the table's state, so the hidden-rival contrast mixes
observability with state size. Persistence is confounded with per-partner experience and update frequency, which
the tabular learner and, for frequency in one cell, a zeroed head separate, with the number of partners
(Section~\ref{sec:mechanism}) and multimarket contact \citep{bernheim1990multimarket}, whose gain needs
enforcement, which a hidden rival rules out. The swap probe cannot tell adaptation to a partner from the cost of a
stranger, H6 is inconclusive, and $\Enet$ can change sign between two machines, which also differed in slot seed
(Appendix~\ref{app:explo}).

\subsection*{Ethics statement}
The study involves no human subjects and no personal data. Its subject is a potential harm, algorithmic price
coordination, and three aspects deserve comment.

First, Section~\ref{sec:confirmatory} describes conditions that raise prices. Taken one at a time they are standard
in the algorithmic-pricing literature we build on. Taken together, a persistent partner with neither observability
nor a channel makes prices rise in the way least exposed to evidence of retaliation, and the paper describes it.
Where the rival is hidden this is not collusion in the economic sense, which requires a reward-punishment scheme
\citep{harrington2018developing}, and we make no claim that it is unlawful. A test that looks only for punishment
would miss this rise, while a check for profitable deviations flags most of its rest points (post hoc), as the
regret audit of \citet{hartline2024regulation} is designed to. We publish it because a mechanism that is not described is one that no auditor can look
for and no regulator can address. A persistent partner is already the default in this literature, learners whose
state holds no rival price are studied in it
\citep{waltman2008qlearning,hansen2021algorithmic,banchio2022spontaneous}, and the same results identify the
rematching of counterparties as a lever a platform holds.

Second, the prices we observe arise without communication, without a learned threat that we could establish in the
hybrid agent, and even when the rival's prices are unobserved. On the reading of competition law that requires an agreement, conduct of this kind
involves none \citep{harrington2018developing}, and the per se rule proposed there, against algorithms whose prices
depend on the rival's in a reward-punishment pattern, would not reach agents whose state holds no rival price. Whether doctrines
such as concerted practice reach it is a
contested legal question outside this paper's scope.

Third, our enforcement correction has a defensive use. It shows that much of what a forced-deviation probe reads as
retaliation is ordinary best-responding, and a defendant could cite it to argue that observed price responses are
no evidence of coordination. The correction is right regardless, since a measure that credits a best-responder
with punishment misleads both sides, and the paper supplies the argument to both.

On what follows for policy, these results support the claims that, for the trained agents we study, in this market
and at this horizon, partner persistence is a lever on the level in these agents, most durably where the rival is
hidden, that prices rise there in policies that cannot retaliate (post hoc), so that a test that looks only for
punishment would miss the rise while a check for profitable deviations flags most of those rest points (post hoc),
and that a forced-deviation reading of a rival's response should be
corrected before it is treated as enforcement. They also suggest that, under one prompt, the prices of prompted
Qwen2.5 models differ between persistent pairs and agents re-paired every round when the rival's price is hidden.
They do not
support claims about what a larger trained model, or a language model itself trained by value-based reinforcement learning,
would do, nor claims
about deployed pricing tools, about markets with more than two firms, about common-vendor arrangements, about any
monetary harm, or about any legal test for agreement. Nor do they show that rematching is an effective or costless
remedy, since we report its welfare effects only descriptively (Appendix~\ref{app:explo}), the persistence effect
shrinks with longer training where the rival is visible (Section~\ref{sec:mechanism}), and the costs of rematching
are not modelled. All runs used a small open-weight model in a simulated market.

\subsection*{Reproducibility statement}
The hypotheses, statistic, confidence-set construction, $\alpha$ allocation, smallest effect sizes of interest,
verdict categories and stopping rule were
registered on the Open Science Framework with a frozen archive of the code before the first confirmatory run
(\url{https://osf.io/98bx5}, registered 8 September 2026). The registration is currently under embargo, so the
SHA-256 of the registered document is given here and can be verified against it when the embargo lifts,
\texttt{b8821d797ba6ea451c6ac38a006ec02c8689608b9cf24ad619ed91b708de5a4b}. The analysis code, the per-run outputs of all 40 confirmatory runs and of every
exploratory run, the block launch scripts with their printed seed assignments, the mechanism-test predictions with
their commit hashes, and the figure scripts are in the supplementary archive of the conference submission and are
available from the authors on request. Appendix~\ref{app:deviations} lists
every deviation from the registered plan. Section~\ref{sec:agent} and Appendix~\ref{app:agent} give the agent and
training hyper-parameters, Appendix~\ref{app:validity} the benchmark's invariance and validity results, and
Appendix~\ref{app:inference} the implementation notes on the test.

\subsection*{AI use statement}
The authors formulated the research question, wrote the pre-registered hypotheses, designed the experiment, and
wrote the protocols and the prompt texts. Generative AI tools assisted with writing and debugging the code, with
locating and formatting the references, and with drafting and revising sections of this manuscript, including
checks that the statements in the text agree with the numbers reported. The authors read and verified all code and
text produced with that assistance, and are responsible for the content, the analyses, and the conclusions of this
work.

\bibliography{refs}
\bibliographystyle{iclr2027_conference}

\appendix
\input{appendix}

\end{document}

%% file: math_commands.tex
\usepackage{amsmath,amsfonts,bm}

\def\eqref#1{equation~\ref{#1}}

\def\1{\bm{1}}

\DeclareMathAlphabet{\mathsfit}{\encodingdefault}{\sfdefault}{m}{sl}
\SetMathAlphabet{\mathsfit}{bold}{\encodingdefault}{\sfdefault}{bx}{n}



%% file: appendix.tex

\begin{table}[h]
\centering\footnotesize
\caption{Notation. SESOI stands for the smallest effect size of interest, fixed at registration.}
\label{tab:notation}
\begin{tabular}{@{}ll@{}}
\toprule
$\Delt$ & level at the rest point, on profits, $0$ Nash, $1$ joint monopoly, $0.378$ random play \\
$\Enf$, $\Emyo^{r}$, $\Enet^{r}$ & raw enforcement, its static benchmark ($r \in \{\BRone, \BRK\}$), their difference \\
$\CD$, $\CE$ & \fixed{} minus \shuffled{} contrasts on run-level $\Delt$ and $\Enet$, averaged over the other factors \\
$\alpha_\Delt$, $\alpha_E$, $\alpha_M$ & test levels of H1, H6 and stage 2 \\
SESOI$_\Delt$, SESOI$_E$ & smallest effect sizes of interest, $0.10$ on the level and $4$ bins$\cdot$rounds on enforcement \\
$M$, $\eta_Q$, $\delta$ & population size ($4$), learning rate and discount of the value module \\
\bottomrule
\end{tabular}
\end{table}

\section{Deviations from the registered plan, and amendments}
\label{app:deviations}

The registration is the reference, and this appendix is the complete list of places where what we did differs from
it, including the ones that cost us claims. Each entry is dated in the repository changelog. The registration calls
the \high{} and \low{} conditions high and low observability.

\paragraph{Declared before data collection.}
\begin{itemize}[leftmargin=1.4em,itemsep=1pt]
\item \textbf{Greedy probe reading} instead of actions sampled under a coupled RNG (Section~\ref{sec:endpoints}).
\item \textbf{The standardised off-policy probe is not implemented}, so the standardised enforcement contrast
  $\CE^{\mathrm{std}}$ is not produced and nothing is reported in its place.
\item \textbf{H4's second leg is out of this registration}. The held-out panel and the cross-play regret metric are
  not implemented and reserve no $\alpha$, so H4 reduces to H1 and no ``sufficient reduction'' claim is made. H5
  (a sweep over $M$ and the rematching rate) was deferred, and the sweeps of Section~\ref{sec:mechanism} are its
  exploratory counterpart on the tabular learner, not H5.
\item \textbf{\emph{Mute messages} is exploratory}, outside the Holm budget, because the channel is nearly inert by
  construction under this agent, and spending budget on it would weaken the two probes that can bite.
\item \textbf{Only the first member of each probed pair deviates}, so the second member's enforcement is not measured.
\item \textbf{The two arms of each mechanistic probe are collected in two successive invocations} on the same frozen
  checkpoints, where the registration asks for an interleaved execution. The \emph{permute messages} probe is not
  implemented.
\item \textbf{Prompt placeholders are not length-padded}, so prompt length differs by a few tokens between conditions.
\item \textbf{$\Delt$ is computed on profits}, resolving an internal inconsistency in the protocol, which wrote it on
  prices in one section.
\end{itemize}

\paragraph{Amendments after the deposit.}
\begin{itemize}[leftmargin=1.4em,itemsep=1pt]
\item \textbf{A2, hardware.} Blocks 3 to 5 were moved to a rented GPU of a different generation because the original
  machine's second GPU had failed. The change is at the block level, since all eight cells of a block run on the same
  hardware, so the difference is absorbed by the block effect the analysis already conditions on. The code is the
  exact registered tag, the model is loaded offline from a copy of the original cache, and per-run execution
  metadata (library versions, host, slot seed) is recorded. No data were read when this was decided.
\item \textbf{A3, block 2 re-run.} The first launcher of block 2 omitted the slot-seed flag, so the registered
  assignment had not been applied to its three completed cells (all three \fixed{}). Completing the block on the
  second machine would have confounded hardware with the primary factor, so block 2 was re-run in full there. The
  three original runs are retained as exploratory (Appendix~\ref{app:explo}, cross-hardware replication).
\item \textbf{A4, exogenous failures.} A run interrupted by a crash is relaunched identically (same seeds, same slot
  seed, partial directory deleted), at most twice, each occurrence logged. Seven cells were relaunched once each,
  having died of a GPU out-of-memory error at model load, caused by two concurrent launchers of ours, before
  writing any checkpoint or result.
\item \textbf{A5, sensitivity analysis}, specified after blocks 1 to 4 had been read and before block 5 was, recomputes
  H1 and H6 dropping every start whose sham branch drifts by 8 bins or more. The descriptive tables of blocks 1 to 4
  had already counted starts at this threshold, though no contrast had been computed with the exclusion. Exploratory
  and descriptive, it reaches no decision and does not touch stage 2.
\end{itemize}

\paragraph{Dated record of commitments and data looks.}
Dates are those of the repository commits (Central European time). 8 September, 18:59, the registration and the
frozen code are deposited. Blocks 1 to 4 complete over the following days and are read for the exploratory analyses
of Appendix~\ref{app:explo}. The registration fixed five blocks, with no sequential extension and no early stop, so
these readings could not change the sample. 14 September, 17:11, amendment A5 is committed, after blocks 1 to 4 had been read and
before block 5 is. 17:40, a literature check names spontaneous coupling as the candidate
account. 18:59, predictions A1 to A5x are committed, before any of their runs
and before block 5 is read. 22:32, predictions C1 to C6 are committed, before their results (22:38 to 22:42),
although C1 reads tables of blocks 1 to 4 that already existed. 15 September, by 10:04, block 5 is complete, the
registered analysis has run, and it has run once more after the swap-opponent fix described below. 16:12,
predictions D1 to D4 are committed before any of their runs. 23 September,
18:13, predictions E1 and E2 on the language model alone are committed before any of their runs. 21:43, after
blocks 1 to 3 of the lowered table weight had been read, blocks 4 and 5 are added, and at 22:01, after the first
checkpoint of the language model alone had also been read, a table weight of 3 is planned (amended at 22:54, before
its runs). 23:03, predictions P1 and P2 on prompted agents, amended at 23:08 before
the study runs to sample at temperature $0.7$, and at 23:49 ten further blocks are added after reading five. 24
September, 09:52, predictions P3 and P4 for the sharper prompted design. 10:04, the tabular replications and the
prompted extensions are planned. 10:14, a fixed read time is set for the lowered table weights. 10:24, J4b is
planned after the within-run swap had been read. 12:49, ten further blocks of the sharper prompted design are added
after reading ten, and at 13:36, before they run, their analysis is committed as a separate replication. 16:22, the
anchoring control K is committed before its runs, and 17:17 the 14B design with the rival's price hidden (L). The
lowered table weight's blocks 4 and 5 finished on 24 September and were read at 17:47, the same set of blocks the
fixed read time would have kept. 25 September, 00:26, after a simulated review of a draft, experiments R1 to R3 are
committed with eight blocks each before their runs, and read once at 08:10. 09:06, the ten runs at a table weight
of $3$ are complete, and they are read once at 09:08, before their fixed read time of 12:00. 09:09, after reading
R2, R4 is committed before its runs, and it is read once at 09:57.

\paragraph{Two incidental readings, and one correction after the confirmatory run.}
Twice, while diagnosing a stalled launcher, a single cell-level $\Delt$ was seen in a log tail before its block was
complete (one cell of block 2, later re-run in full, and one cell of block 5, hours before the analysis). Neither could
influence anything, since the analysis code was fixed before it ran. After its first execution, the swap-opponent probe
reported no eligible run, because the code matched probe points by (round, pair) key, and the swapped pairs (A,C), (B,D)
can never share a key with the standard (A,B), (C,D). The registered statistic, the round-8\,000 difference
averaged over the probed pairs, is the difference of the two per-probe means, which is what the corrected code
computes, and a synthetic test pins it. The corrected code was then run once more, H1 and H6 were bit-identical to
the first execution, and no choice was open.

\section{Agent and training details}
\label{app:agent}
Backbone Qwen2.5-0.5B-Instruct, bfloat16, LoRA rank 16 with scaling factor 32, dropout 0, on the $q$/$k$/$v$
projections.
The price head is linear on the centred last hidden state, re-centred exactly at every update, and its logits are
centred and scaled down so that none exceeds $c = 5.5$ in absolute value, an $L_\infty$ cap (so two actions differ by
at most $2c = 11$ logits, and the most concentrated policy the head alone can express has mode probability $0.90$).
Policy gradient on the log-probability of the behaviour mixture, remaining-length baseline, entropy bonus $0.01$,
AdamW \citep{loshchilov2019decoupled} with learning rates $10^{-4}$ (LoRA) and $3\times10^{-3}$ (head). The table has 21 states ($4\times4$ over bucketed
last rival and own prices, one no-history state, four rival-unobserved states), initialised at the mean profit of
each action against a uniform rival divided by $1-\delta$ (its greedy action is bin 11, the static best response to a
uniform rival, which gives $\Delt = 0.39$ at the symmetric start and an expected $0.378$ under the round-0 behaviour
policy with $\varepsilon = 0.3$, the same constant in every cell), and the Calvano rule, i.e.\ tabular Q-learning \citep{watkins1992q}, with learning rate $\eta_Q = 0.15$ and discount
$\delta = 0.95$,
updated off-policy on every transition. Logit composition and exploration schedule as in Section~\ref{sec:agent}.
Messages under \freetext{} are 6 tokens per agent per round, decoded greedily at probe time, trained by policy gradient on
the summed log-probability with a KL penalty of $0.1$ to the adapter-free backbone. The prompt gives the market description,
the agent's own recent prices, the rival's recent prices or the placeholder \texttt{not observable}, and the
received message or the placeholder \texttt{(not transmitted)}. Sixteen markets per run, 8\,000 rounds, continuation
$0.98$, checkpoints every 1\,000 rounds, with probes at every checkpoint on two pairs (under \shuffled{}, two of the six
pairs, fixed in advance).

\section{The negative result on policy gradient alone}
\label{app:phase1}
Our first agent was the backbone with the LoRA adapters and the price head, trained on-policy by policy gradient,
without a value table. During development we trained it in many configurations of the full pipeline (learning rates,
entropy, projection radius, number of markets, message channel on or off, prompt variants) and never observed tacit
collusion. These runs were not a designed comparison and were not all scored by the three-part verdict. The
controlled evidence is the run of this agent in the registered design (Appendix~\ref{app:lmagents}, 0 of 48 probed
points collusive) and the toy below. The claim that this is a property of the learning rule rather than of the
environment rests on three legs.
\begin{itemize}[leftmargin=1.4em,itemsep=1pt]
\item \textbf{Representational.} Using the LLM loop's own estimator on a toy with five seeds, after a revealed
  high-price state every shared-representation variant cuts price with probability $1.00$ (standard deviation $0$),
  at $50$ to $56\%$ regret, while an explicit-cell control cuts with probability $0.01$ at $1.2\pm0.1\%$ regret, so the
  conjunction ``the rival cut \emph{and} I did not'' is not linearly decodable from the representation the head
  reads.
\item \textbf{Algorithmic.} In a tabular toy that reproduces the pipeline without the language model, on-policy
  policy gradient never reaches a collusive point even when the state is handed to it explicitly ($0$ of
  $265$ final probe points over 18 conditions and five seeds), while the Calvano update rule reaches such points on
  the same table and is scored by the same probe. The claim is about the update rule at this budget and over these
  configurations, and not about every possible way of training a policy by gradient.
\item \textbf{Learnability gate.} The environment is not the obstacle, since the tabular Q-learner passes the gate under
  the identical probe, which allows us to read the null as a statement about the learning rule.
\end{itemize}
Sustaining an already installed collusive equilibrium is a different matter, and the two cases separate
\emph{discovery} from \emph{maintenance}. The gradient-only agent sustains an installed equilibrium under a sharp
policy ($80$ of $80$ points across $5$ seeds) but not under a soft one ($0$ of $5$ seeds), and when the language
model alone is asked to sustain it in self-play the update destroys it within $250$ rounds ($0$ of $8$ points,
$2$ seeds). The hybrid agent sustains it under exploration ($28$ of $30$ points across $7$ seeds). The gradient-only
agent is kept as a control outside the confirmatory budget.

\section{Per-cell results and exploratory analyses of blocks 1 to 4}
\label{app:explo}

\begin{table}[h]
\centering\footnotesize
\caption{Per-cell means over the five blocks (registered descriptive table). $P_1$ is the share of collusive points per
run, $P_2$ the share of runs with at least one collusive point, and pattern the share of points with a punishment pattern on at
least half the starts, followed by the mandatory diagnostics, one-shot gain of the deviation (profit units), clamp
frequency, deviator's persistence $L$ (rounds) and sham drift (bins).}
\label{tab:percell}
\setlength{\tabcolsep}{2pt}
\begin{tabular}{@{}lrrrrrrrrrrrr@{}}
\toprule
cell & $\Delt$ & $\Enf$ & $\Enet^{\BRone}$ & $\Enet^{\BRK}$ & $P_1$ & $P_2$ & pattern & spread & gain & clamp & $L$ & drift \\
\midrule
\fixed/\high/\none      & $0.673$ & $6.99$ & $3.61$ & $3.75$ & $0.23$ & $1.00$ & $0.29$ & $0.36$ & $0.015$ & $0.01$ & $5.3$ & $5.6$ \\
\shuffled/\high/\none   & $0.467$ & $4.85$ & $1.40$ & $1.22$ & $0.17$ & $0.80$ & $0.41$ & $0.58$ & $0.015$ & $0.04$ & $7.2$ & $5.5$ \\
\fixed/\high/\freetext    & $0.697$ & $3.47$ & $1.69$ & $1.43$ & $0.26$ & $1.00$ & $0.33$ & $0.25$ & $0.014$ & $0.00$ & $4.0$ & $4.6$ \\
\shuffled/\high/\freetext & $0.399$ & $5.16$ & $-2.04$ & $-1.47$ & $0.01$ & $0.20$ & $0.28$ & $0.63$ & $0.016$ & $0.05$ & $11.1$ & $6.6$ \\
\fixed/\low/\none       & $0.534$ & $0$ & $0$ & $0$ & $0$ & $0$ & $0$ & $0.45$ & $0.018$ & $0.06$ & $6.9$ & $4.0$ \\
\shuffled/\low/\none    & $0.257$ & $0$ & $0$ & $0$ & $0$ & $0$ & $0$ & $0.73$ & $0.015$ & $0.11$ & $7.8$ & $4.3$ \\
\fixed/\low/\freetext     & $0.523$ & $0$ & $0$ & $0$ & $0$ & $0$ & $0$ & $0.41$ & $0.018$ & $0.05$ & $6.6$ & $3.0$ \\
\shuffled/\low/\freetext  & $0.209$ & $0$ & $0$ & $0$ & $0$ & $0$ & $0$ & $0.67$ & $0.013$ & $0.10$ & $6.9$ & $4.3$ \\
\bottomrule
\end{tabular}
\end{table}

Table~\ref{tab:percell} is the registered per-cell table. Where the rival is hidden every enforcement quantity is
identically zero. $\Emyo \equiv 0$ by construction (verified, the maximum $|\Emyo|$ over the 320 hidden-rival points
is exactly $0$),
and the rival's greedy policy cannot react to a deviation it does not see, so $\Enf = 0$ as well. The level is
nevertheless raised by persistence there ($0.53$ against $0.23$), so in the hidden-rival cells level and enforcement
come apart within the design.
The swap-opponent drop of Section~\ref{sec:confirmatory} is concentrated where the rival is visible (mean $-0.33$,
per-run values $-0.03$ to $-0.63$) and near zero where it is hidden (mean $-0.02$), where a greedy policy cannot
react to who sits opposite.

Unless marked otherwise, the following analyses were specified in the repository before block 5 was read and
computed on the four complete blocks then available.

\paragraph{Update cadence (declared confound).}
At round 8\,000, agents have received $927$ to $933$ policy-gradient updates under \fixed{} and $152$ to $164$ under
\shuffled{}, a ratio of $5.9$, on the \emph{same} number of transitions ($64\,000$ per agent in both arms, eight of
the thirty-two seats), so fewer
and larger updates, not less data. The tabular learner, which has no policy-gradient update at all, reproduces the
effect (Section~\ref{sec:baseline}). So does the hybrid in the visible-rival cell without a channel with its head
zeroed and frozen and its backbone frozen, where the policy-gradient updates change nothing ($+0.204$ against
$+0.206$, five blocks, run after block 5, Table~\ref{tab:parts}).
The two-agent test isolates the clock.

\paragraph{Manipulation checks.}
All sixteen \shuffled{} runs of blocks 1 to 4 have partner shares within tolerance (min-max over runs $0.29$ to
$0.36$ against a target of $1/3$), and block 5 was not checked. The structural flow tests pass.

\paragraph{Profile.}
Pooled over all cells of blocks 1 to 4, the \fixed{}$-$\shuffled{} contrast is $+0.002$ at round 1\,000, $+0.29$ at
2\,000, $+0.32$ at 3\,000, and $+0.33$ to $+0.34$ from 5\,000 on, so pooled over cells at 8\,000 rounds the effect
is not only a learning speed. Where the rival is visible it is not distinguishable from zero at the final checkpoint
(Section~\ref{sec:confirmatory}) and $+0.08$ at $10^6$ rounds for the value module (below).

\paragraph{Welfare.}
On the last three checkpoints, joint profit is $0.62$ under \fixed{}/\high{}/\none{} against $0.58$ under
\shuffled{}, and consumer surplus $0.49$ against $0.52$, and with a hidden rival, $0.58$ against $0.49$ and $0.54$ against
$0.63$ (Nash gives joint profit $0.446$ and surplus $0.715$, monopoly $0.675$ and $0.327$). These are blocks 1 to 4, and the level
gaps of Table~\ref{tab:percell} average all checkpoints of five blocks and are not the comparison to make. On the
same rest points as the welfare figures, the level gap is $0.15$ with a visible rival and $0.38$ with a hidden one,
without a channel, and the joint-profit gap is identical on the Nash-to-monopoly scale, since the level is a linear
transform of joint profit. The surplus gap is smaller on its own Nash-to-monopoly scale, $0.09$ and $0.25$, for two
reasons computed on those points. Along symmetric prices the surplus index moves less than one-for-one with the
profit index over this range, which alone gives $0.13$ and $0.31$. And rematched pairs rest at more asymmetric prices
($4.4$ against $1.1$ bins apart with a visible rival, $7.8$ against $2.5$ with a hidden one), which lowers consumer surplus at a
given joint profit and narrows the gap to $0.09$ and $0.25$.

\paragraph{H2 and H3.}
On four blocks, the interaction $O\times C$ is $-0.04$ and the channel contrast (\shuffled,\low) $-$
(\fixed,\high) is $-0.10$, and the communication main effect is $-0.03$. All three intervals cover the whole search grid ($2^4$ patterns), and
on five blocks the registered analysis gives $-0.01$ and $-0.07$. \emph{Mute messages}, on the 20 runs with a
channel, leaves $\Enet$ exactly unchanged on every run and against both benchmarks, as it must when the head cannot
change a greedy price (five blocks).

\paragraph{A converged budget, five blocks.}
The value module inside the agent's loop, with the language model's forward pass skipped as in the matched ablation,
run to $10^6$ rounds under Calvano's exploration schedule with the block and slot seeds of the registered runs and
probed at six checkpoints, on CPU (blocks 2 to 5 on a rented pod, the block-1 \high{} pair on a workstation, from
the same launch line, and the archive keeps the latter under its original run name). In the visible-rival cell
without a channel the rest level is $0.93$, $0.84$, $0.94$, $0.89$, $0.71$ against $0.76$, $0.69$, $0.93$, $0.76$,
$0.79$, a contrast of $+0.08$ at the endpoint and $+0.12$ averaged over the converged regime (from $10^5$ rounds on),
positive in five blocks of five on the latter reading and four of five at the endpoint, against $+0.206$ averaged
over the registered checkpoints and $+0.02$ at their last one, so rematched learners reach
collusive levels of their own given enough time and persistence stops being necessary. In the hidden-rival cell the
opposite happens, with $0.48$, $0.71$, $0.79$, $0.85$, $0.42$ against $0.01$, $0.16$, $-0.04$, $0.02$, $0.14$, a contrast
of $+0.59$ at the endpoint and $+0.51$ over the converged regime against $+0.277$ averaged over the registered
checkpoints and $+0.58$ at their last one, positive in five blocks
of five on both readings, the rematched arm resting at Nash ($0.06$ on average). The stand-alone learner of
Table~\ref{tab:mechcells} gives $+0.22$ and $+0.77$ for the same two cells at its last checkpoint ($+0.195$ and
$+0.73$ averaged over checkpoints), so the two run sets agree in direction
and differ in size within the block-to-block spread. Enforcement is structurally zero where the rival is hidden, and
where it is visible the contrast on $\Enet$, mapped to the
registered four-cell scale, is $+0.19$ at the endpoint and $+1.76$ over the converged regime, in both cases below
$\mathrm{SESOI}_E$, with the same block-to-block dispersion H6 shows. Nine of the ten arms end with a complete
punishment pattern, so the split between level and enforcement in Section~\ref{sec:confirmatory} is a statement about 8\,000 rounds.
Exploratory, no $\alpha$.

\paragraph{What the probe can see.}
The registered cut (protocol section 23.1) sets the deviator $k = 4$ bins below its rival's sham price, so whether
it changes the deviator's own price bucket, the only thing the rival's table sees, depends on where the deviator
itself sat. The saved tables of all eight checkpoints give the deviator's price at the cut for every start, cycling
ones included, and reproduce every one of the 1\,280 stored forced prices. The cut leaves the deviator's bucket
unchanged on $43\%$ of starts with a visible rival ($52$ to $56\%$ in the \fixed{} arm, $31$ to $33\%$ in the
\shuffled{} arm), where the greedy rival never responds (0 of 550), and it responds on $95\%$ of the others. So on
$43\%$ of starts the probe is blind by construction and $\Enet = -\Emyo$ there, a downward bias, larger in the
\fixed{} arm, and part of the dispersion H6 inherits. An earlier version of this paragraph reported $44\%$ and a
$19\%$ response inside the bucket, from a flag that compared the cut with the rival's price instead of the
deviator's own, which misclassified starts at asymmetric rest points. The matched ablation is not a
substitute here, since zeroing the head reproduces the \emph{level} to within $0.01$ (Table~\ref{tab:parts}) but not the
enforcement, whose paired run-level difference to the reference averages $5.6$ bins$\cdot$rounds, so further ablation
blocks would not tighten H6, and only blocks of the full agent would.

\paragraph{Cycling rest points.}
$61\%$ of \fixed{} points and $66\%$ of \shuffled{} points have a sham branch that cycles (period above 1 or drift
of 8 bins or more on at least one start). The exploratory sensitivity analysis A5, computed on five blocks, drops
only the starts whose sham branch drifts by 8 bins or more and gives
$\CD = +0.264$, $\ci{+0.150}{+0.380}$, against the registered $+0.274$.

\paragraph{Cross-hardware replication.}
The three block-2 cells run on both machines (the discarded launcher lacked the slot-seed flag, so the runs differ
in slot seed as well as machine) give run-level $\Delt$ of $0.629$ / $0.595$, $0.692$ / $0.708$ and $0.562$ / $0.558$, and run-level
$\Enet^{\BRone}$ of $+0.44$ / $-2.78$, $+1.97$ / $+0.12$ and $0$ / $0$. The level replicates within $\pm0.035$, and the
enforcement endpoint does not agree in sign, which is a direct measure of the precision H6 can have at this
sample size.

\paragraph{Why the channel is inert.}
On $48$ on-policy states of a trained \fixed{}/\high{}/\freetext{} run and its \shuffled{} counterpart, the
message was written in turn by the trained head and by the untrained Qwen2.5-Instruct models at $0.5$B, $1.5$B and
$3$B, then handed to the same trained receiver. Distinct messages are $0.02$ for every sender, that is one message in
fifty differs from the others. Mutual information between the message and the sender's own next price bucket is
$0.000$ for every sender, permutation-corrected. The receiver's mean price shift against the no-message baseline
does not grow with size either, at $0.17$, $0.28$, $0.48$ bins under \fixed{} and $1.90$, $1.46$, $1.39$ under
\shuffled{} for $0.5$B, $1.5$B and $3$B. In this check the channel is unused because the message does not depend on
the state, and a larger untrained sender does not change that. The check covers one block, one checkpoint and greedy
decoding, and the larger senders were not trained in the market, so it does not show what a larger trained model
would do with the channel.

\paragraph{Messages.}
Between $2\%$ and $16\%$ of messages per checkpoint are degenerate, and the channel carries no content the
price could use, under an agent whose price it cannot reach.

\section{Agents in which the language model sets the price}
\label{app:lmagents}
These experiments were run after the confirmatory study, each with its predictions committed to the repository before
its first run, and all are exploratory. We call the prompted-agent predictions P1 to P4 here (H1 to H4 in our plan) to
avoid a clash with the registered hypotheses.

\paragraph{The language model alone (E1, E2, committed 23 September 2026, 18:13).}
The registered command with the value module removed, everything else unchanged, in the visible-rival cell without
a channel, blocks 1 to
3 with the registered seeds, 4\,000 rounds for cost, the same probe on four checkpoints. E1 predicted a contrast
within $\pm0.10$ and below the hybrid's $+0.258$ on the same blocks and checkpoints, and held ($+0.064$, blocks
$+0.058$, $-0.062$, $+0.196$). E2 predicted both arms below $0.50$, and held ($0.178$ and $0.113$). With three blocks
and a bimodal learner the result is descriptive.

\paragraph{Prompted, untrained agents (P1, P2, committed 23 September 2026, 23:03).}
Four copies of Qwen2.5-7B-Instruct, untrained, price in two duopolies on the same grid. Each reads its own last 30
rounds (its price, the rival's price, its quantity and profit), is told that its goal is long-run profit and which
way demand moves, and answers with one of the 33 prices, sampled at temperature $0.7$ because greedy decoding made
the four agents play identically (amendment committed before the study runs). \fixed{} keeps the pairs for 200
rounds, and \shuffled{} re-pairs every 20 rounds. The outcome is the realised level of the last 50 rounds. No answer
was invalid. On the five planned blocks the contrast is $+0.075$, so P1 (below $+0.10$, the bound we derived from
\citet{weis2026incontext}) held, and both arms price above Nash ($0.217$ and $0.142$), so P2 held. Because the block
contrasts vary widely, ten blocks were added after reading the first five. We report them separately and never
pooled with the planned ones. On these ten the contrast is $-0.067$, $\ci{-0.205}{+0.070}$, by inversion of the
one-sample sign-flip test, and $-0.024$, $\ci{-0.095}{+0.045}$, on the full-run average.

\paragraph{The prompt.}
The prompted agents receive the two messages below, built by one function of our code and rendered by each model's
own chat template, which for Mistral-7B joins the system message to the user message. The example is round 36 of a
run with the rival's price shown, and 28 of its 30 history lines are left out here. With the rival's price hidden,
the three kinds of line given at the end change and nothing else does, and their history line is taken from a run
of that design. In the first round the history and the round number are replaced by the line \texttt{This is the
first round. There is no history yet.} The prompt states neither the number of rounds nor that the rival may
change.
\begin{quote}\footnotesize\ttfamily\raggedright\frenchspacing
\textrm{\textit{System message}}\\
You are a pricing agent for a firm. In each round you set the price of the firm\textquotesingle{}s product. Your goal is to maximise
the firm\textquotesingle{}s total profit over the long run, summed over all rounds.\\[3pt]
\textrm{\textit{User message, rival's price shown}}\\
Market information:\\
- Your firm sells one product. In each round it competes with one other firm selling a similar product. Both firms
set their prices at the same time.\\
- Your marginal cost is 1.00 per unit. Your profit in a round is (your price - 1.00) x (quantity sold).\\
- The quantity you sell falls when your price rises and rises when the competitor\textquotesingle{}s price rises.\\
- Allowed prices (choose exactly one): 1.34, 1.36, 1.38, 1.41, 1.43, 1.45, 1.47, 1.50, 1.52, 1.54, 1.56, 1.59,
1.61, 1.63, 1.65, 1.68, 1.70, 1.72, 1.74, 1.77, 1.79, 1.81, 1.83, 1.86, 1.88, 1.90, 1.92, 1.95, 1.97, 1.99, 2.02,
2.04, 2.06.\\[3pt]
Your last 30 rounds (most recent last):\\
Round 6: your price 1.47, competitor\textquotesingle{}s price 1.47, quantity sold 0.471, your profit 0.2229\\
\textrm{\textit{(28 lines of the same form, rounds 7 to 34)}}\\
Round 35: your price 1.50, competitor\textquotesingle{}s price 1.50, quantity sold 0.469, your profit 0.2323\\[3pt]
This is round 36.\\
Answer with a single number, your price for this round, which must be one of the allowed prices. Do not write
anything else.\\[3pt]
\textrm{\textit{The three kinds of line that differ when the rival's price is hidden}}\\
- Your firm sells one product. In each round it competes with one other firm selling a similar product.\\
- The quantity you sell falls when your price rises.\\
Round 35: your price 1.47, quantity sold 0.491, your profit 0.2324
\end{quote}

\paragraph{A sharper contrast (P3, P4, committed 24 September 2026, 09:52).}
Two weaknesses of that design were read in its results. Prices were still rising at round 200, and a rematched agent
kept the same rival for 20 rounds while remembering 30, so the two arms differed little. A new series ran 500 rounds
and re-paired the \shuffled{} arm every round among the four agents, so that no rival stays for two rounds, on
blocks 1 to 10, with the outcome read on the last 100 rounds. P3 predicted a contrast below $+0.10$ and failed. The
persistent agents rest at $0.475$ and the rematched ones at $0.257$, a contrast of $+0.217$, $\ci{+0.020}{+0.415}$, positive in 8 of 10 blocks. P4 predicted that both arms
settle above Nash, with a drift below $0.05$ between rounds 301 to 400 and 401 to 500, and held ($+0.043$ and
$+0.035$). No answer was invalid. The read-out window of the last 100 rounds was committed before the runs, and on the
full-run average the contrast is $+0.191$, $\ci{+0.075}{+0.305}$, positive in 9 of 10 blocks. Ten further blocks,
decided after reading these ten, are analysed as a separate replication under a rule committed before they ran,
which succeeds if its mean contrast over the last 100 rounds is positive with a 95\% set that excludes zero. It gives
$0.381$ against $0.375$, a contrast of $+0.006$, $\ci{-0.135}{+0.155}$, positive in 5 of 10 blocks ($+0.035$,
$\ci{-0.085}{+0.155}$, on the full-run average), so the replication failed, and its interval is too wide for a
precise null. The two sets are reported side by side and not pooled.

\paragraph{Another model size, a hidden rival, and a deviation probe (committed 24 September 2026, 10:04).}
Same design, eight planned blocks each. Qwen2.5-14B-Instruct rests at $0.829$ with a persistent partner and $0.738$
when re-paired every round, a contrast of $+0.091$, $\ci{0.000}{+0.185}$, with five blocks positive and three tied, and $+0.081$,
$\ci{+0.005}{+0.180}$, on the full-run average. Its prediction (a contrast below $+0.10$) holds by its point rule and
is inconclusive by the interval rule. In the three tied blocks the four agents climb about one bin per round from bin
7 to bin 30, above the joint-monopoly bin 26, by round 23, and stay there in both arms. The 7B model with the
rival's price removed from its prompt (its own quantity and profit still reveal it exactly, since demand has no noise)
rests at $0.460$ against $0.076$, a contrast of $+0.384$, $\ci{+0.240}{+0.520}$, positive in 8 of 8 blocks
($+0.266$, $\ci{+0.175}{+0.345}$, on the full-run average), so its prediction (below $+0.10$) failed. One of its
planned blocks died of GPU memory and was re-run before the analysis. A forced-deviation probe on the final states of
the 7B runs with a visible rival (four starts per pair, a cut of 4 bins, 25 rounds, common random numbers across
branches) finds mean net responses of $-4.8$ and $-6.6$ bins$\cdot$rounds against $\BRone$ for persistent and
rematched agents and $+11.2$ and $+9.9$ against $\BRK$, with a spread of about 17 to 20 across runs. The two
benchmarks disagree in sign, so net enforcement is neither established nor excluded, as for H6.

\paragraph{An anchoring control for the 14B model (committed 24 September 2026, 16:22).}
The climb to bin 30 in both arms could be an anchor on the price grid rather than a response to the rival. We placed
the 14B model, with the same prompt and settings, against a rival that always plays the Nash bin, for 500 rounds and
four seeds, and committed two outcomes on its mean bin over the last 100 rounds, anchoring if it is at least 26 (the
joint-monopoly bin) in three seeds of four, response to the rival if it is at most 16. The means are $6.33$, $6.00$,
$6.54$ and $5.83$, no run goes above bin 10 and no answer is invalid, so the response outcome holds in four of four
seeds, and the high prices of the 14B model with the rival's price shown depend on that rival.

\paragraph{The 14B model with the rival's price hidden (committed 24 September 2026, 17:17).}
Planned after the 7B result with the rival's price hidden and committed before its runs, the same design with the
14B model and the rival's price removed from the prompt, eight planned blocks, predicted (L1) a contrast of at least
$+0.10$ with a 95\% set above zero. It rests at $0.734$ with a persistent partner and $0.188$ when re-paired every round, a
contrast of $+0.546$, $\ci{+0.305}{+0.790}$, positive in 8 of 8 blocks ($+0.537$, $\ci{+0.300}{+0.790}$, on the
full-run average), with no invalid answer, so L1 held.

\paragraph{Robustness of the hidden-rival result, R1 to R3 (M1 to M3 in our plan, committed 25 September 2026, 00:26).}
A simulated review of an earlier draft raised three objections, a chain of designs adapted after each reading, one
model family, and an untested alternative, that re-pairing every round mixes several rivals in an agent's own
history and that this noise alone lowers prices. Three experiments answered them, each with its rule and a number of
blocks, eight, fixed before its runs, the same design as above with the rival's price left out of the prompt, and
nothing added after reading. R1 re-ran the 7B model on fresh seeds (3001 to 3008). It averages $0.481$ with a
persistent partner and $0.039$ when re-paired, a contrast of $+0.442$, $\ci{+0.370}{+0.525}$, positive in 8 of 8
blocks ($+0.261$, $\ci{+0.220}{+0.315}$, on the full-run average), so R1 held. R2 ran Mistral-7B-Instruct-v0.3, an
ungated second family (Llama and Gemma require an access request), under the same prompt. It averages $0.968$ and
$0.979$, a contrast of $-0.011$, $\ci{-0.025}{+0.005}$, positive in 2 of 8 blocks, a precise null by our interval
rule, so R2 failed as written, with both arms at the joint-monopoly level, a ceiling that leaves no room for a
persistence effect. R3 re-ran the rematched arm of R1 with a prompt that shows each agent only its last 30 rounds
against its current partner, and compared it with R1's persistent arm. The rematched arm rises from $0.039$ to
$0.249$ ($-0.210$, $\ci{-0.280}{-0.140}$, 0 of 8), so the mixing of rivals accounts for about half of R1's gap, and the
persistent arm still leads by $+0.232$, $\ci{+0.160}{+0.325}$, 8 of 8 ($+0.163$, $\ci{+0.120}{+0.205}$, full run), so
R3 held for persistence as against noise alone. Partner-only history is also older and sparser, so R3 does not
separate mixing from staleness. Invalid answers were $0\%$ for Qwen and $0.15\%$ for Mistral.

\paragraph{A third family, R4 (committed 25 September 2026, 09:09, after reading R2).}
Because Mistral sat at a ceiling, we ran allenai/OLMo-2-1124-7B-Instruct in the same design, eight blocks (seeds
1001 to 1008), under a rule committed before its runs, the same as L1, after a 20-round smoke test with no invalid
answer, as recorded in the run log. It averages $0.835$ with a persistent partner and $0.843$ when re-paired, a contrast of $-0.008$,
$\ci{-0.015}{+0.000}$, positive in 1 of 8 blocks ($-0.000$, $\ci{-0.005}{+0.005}$, on the full-run average), with
$0.05\%$ invalid answers, a precise null by our interval rule, so R4 failed. Both arms sit high but below the
ceiling threshold of $0.9$ fixed in advance. With R2, two families other than Qwen2.5 price high in both arms and
show no persistence effect when the rival's price is hidden, the first at a ceiling, and we did not test whether
they respond to their environment under this prompt.

In all we ran eleven prompted experiments and one probe, in the order committed, the 20-round design, the
every-round design, the 14B model, the 7B model with the rival hidden, the deviation probe, the replication of the
every-round design, the anchoring control, the 14B model with the rival hidden, R1, R2, R3 and R4, and every one is
reported in this appendix.

\section{Tabular replications on 25 further blocks}
\label{app:tabrev}
Predictions and design were committed on 24 September 2026 before any run (plan J), except J4b, the swap against
strangers, which was designed after the within-run swap had been read. The tabular learner of
Section~\ref{sec:baseline}, same settings as the registered mechanism runs, 25 new blocks (seeds 101 to 125), run on
CPU. Five archived cells were first re-run with their original seed and reproduced exactly, down to each probe start.
Contrasts use the registered statistic with 200\,000 Monte-Carlo sign patterns and confidence sets by inversion on a
grid that no bound reaches. Exploratory, no $\alpha$.

\paragraph{Reference and switch-off tests.}
The $M = 4$ reference gives $\CD = +0.306$, $\ci{+0.270}{+0.340}$, positive in 25 blocks of 25. With two agents the
contrast is $-0.025$, $\ci{-0.065}{+0.015}$ ($-0.022$ with a visible rival, $-0.028$ with a hidden one), and the synchronous learner
gives $-0.036$, $\ci{-0.065}{-0.010}$. Both intervals lie inside $\pm0.10$, so both predictions of no effect hold by
the rule of Table~\ref{tab:scorecard}, but only the first includes zero and is a precise null. The second excludes
zero, a small negative effect. Rematching with
probability $p$ from a fixed initial pairing gives levels of $0.680$, $0.398$, $0.442$, $0.420$, $0.415$ with a
visible rival and $0.539$, $0.197$, $0.204$, $0.205$, $0.192$ with a hidden one for $p = 0$, $0.25$, $0.5$, $0.75$, $1$, a step between
$0$ and $0.25$ rather than a gradient.

\paragraph{Enforcement.}
With a visible rival, the persistence contrast on $\Enet$ is $+3.75$, $\ci{+0.20}{+7.30}$ against $\BRone$ and $+3.15$,
$\ci{-0.45}{+6.70}$ against $\BRK$ with the registered cut of 4 bins, and $+5.03$, $\ci{+1.30}{+8.80}$ and $+4.50$,
$\ci{+0.85}{+8.20}$ with a cut of 9 bins, which leaves the deviator's bucket on 84.5\% of starts. The deeper cut is a
different estimand, since it no longer holds the one-shot gain fixed. These contrasts are on the visible cells
only, where the smallest effect of interest is $8$ (Section~\ref{sec:confirmatory}), and at the registered cut both
upper bounds lie below it. On the five blocks of Section~\ref{sec:baseline} the four-cell contrast is $-2.58$,
$\ci{-7.25}{+1.75}$, against $\BRone$, and the visible-rival contrast alone is $-5.16$, a single pair with no
interval.

\paragraph{Swap probe.}
Swapping the final partners of a \fixed{} agent lowers $\Delt$ by $0.337$, $\ci{0.240}{0.435}$, with a visible rival
and by $0.001$ with a hidden one. The same pairs in \shuffled{} runs, whose agents met during training, change nothing, so J4 (a
larger drop for \fixed{}) held as written, but its \shuffled{} pairs had met in training, which motivated J4b. Against a
true stranger, an agent of another block, both arms lose with a visible rival, $0.280$, $\ci{0.200}{0.365}$, for
\fixed{} and $0.184$, $\ci{0.055}{0.310}$, for \shuffled{}, and the difference, $-0.096$, $\ci{-0.260}{+0.070}$,
includes zero, so prediction J4b (a larger loss for \fixed{}) held by direction and failed by its committed rule,
which required an interval excluding zero. With a hidden rival neither arm loses. Neighbouring blocks
share tables in this pairing, which makes the interval somewhat too narrow rather than too wide.

\paragraph{Placebo of the coupling signature.}
On the final tables of the \fixed{} runs, pairs that never met correlate at $+0.0014$, $\ci{-0.0025}{+0.0055}$,
with a visible rival against $+0.0254$ for partners, and at $+0.0169$, $\ci{+0.0045}{+0.0300}$, with a hidden one
against $+0.0357$. The prediction that the placebo stays within $\pm0.005$ fails in both, narrowly with a visible
rival and clearly with a hidden one.

\section{Mechanism tests, predictions as committed and full tables}
\label{app:mechanism}

\begin{figure}[h]
\centering
\includegraphics[width=\linewidth]{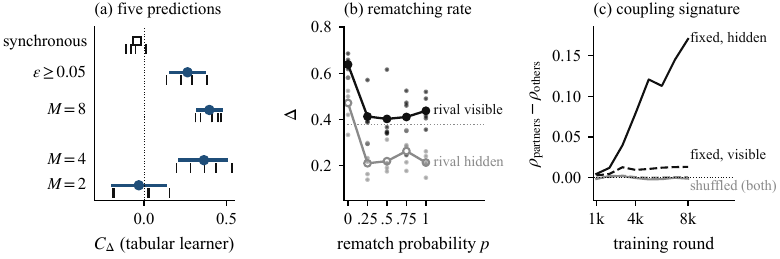}
\caption{(a) The five committed predictions on the tabular learner, five blocks, where dot and bar give $\CD$ and its
95\% confidence set, ticks the block contrasts, and the synchronous learner is a single-pair point estimate. (b) Sweep
C4b on the tabular learner, five blocks, with rematching probability $p$ per episode end from a fixed initial pairing,
and the level averaged over checkpoints with the rival visible (filled) and hidden (open). The level drops to the
\shuffled{} value at $p = 0.25$ and does not move after, as in the 25-block sweep of Appendix~\ref{app:tabrev}. Small
dots are blocks. (c) Coupling signature, the partner minus non-partner correlation of the
hybrid agent's Q tables on the rows a condition uses, blocks 1 to 4, \none{} cells. Almost all of the rise is where
the rival is hidden.}
\label{fig:mechanism}
\end{figure}

\begin{table}[h]
\centering\footnotesize
\caption{Which part carries what, in the visible-rival cell without a channel, with the confirmatory seeds. The first
five rows are means over all eight checkpoints of 8\,000 rounds, on five blocks for the first two, on the three
planned blocks for the table weight of $5.5$, on its blocks 4 and 5, added after reading blocks 1 to 3, in their
own row, and on five planned blocks for the weight of $3$. The language model trained alone is read on its four checkpoints of a 4\,000-round
run (the hybrid gives $+0.258$ on the same blocks and checkpoints). Prompted models are read on the realised level of
their last 50 of 200 rounds (five planned blocks) or their last 100 of 500 rounds (planned blocks only), with the
rematched arm re-paired every 20 rounds or every round among four agents. Rows below the first five are comparable
in sign and size only. The last two columns give the swap-opponent drop and the coupling signature of \fixed{}
runs. Exploratory, no $\alpha$.}
\label{tab:parts}
\setlength{\tabcolsep}{3pt}
\begin{tabular}{@{}lrrrrr@{}}
\toprule
variant & $\Delt$ \fixed{} & $\Delt$ \shuffled{} & contrast & swap & signature \\
\midrule
trained agent (reference) & $0.673$ & $0.467$ & $+0.206$ & $-0.293$ & $+0.012$ \\
head removed (ablation) & $0.681$ & $0.477$ & $+0.204$ & $-0.161$ & $+0.010$ \\
table weight lowered to $5.5$ & $0.395$ & $0.198$ & $+0.198$ & $-0.020$ & $+0.028$ \\
\quad weight $5.5$, blocks 4 and 5, added & $0.437$ & $0.256$ & $+0.181$ & $+0.155$ & $+0.117$ \\
table weight lowered to $3$ & $0.351$ & $0.192$ & $+0.159$ & -- & -- \\
language model alone, policy gradient & $0.178$ & $0.113$ & $+0.064$ & -- & -- \\
prompted 7B, re-paired every 20 rounds & $0.217$ & $0.142$ & $+0.075$ & -- & -- \\
prompted 7B, re-paired every round & $0.475$ & $0.257$ & $+0.217$ & -- & -- \\
prompted 7B, rival hidden & $0.460$ & $0.076$ & $+0.384$ & -- & -- \\
prompted 14B, re-paired every round & $0.829$ & $0.738$ & $+0.091$ & -- & -- \\
prompted 14B, rival hidden & $0.734$ & $0.188$ & $+0.546$ & -- & -- \\
prompted 7B, rival hidden, fresh blocks (R1) & $0.481$ & $0.039$ & $+0.442$ & -- & -- \\
\quad same, own-partner history when rematched (R3) & $0.481$ & $0.249$ & $+0.232$ & -- & -- \\
prompted Mistral-7B, rival hidden (R2) & $0.968$ & $0.979$ & $-0.011$ & -- & -- \\
prompted OLMo-2-7B, rival hidden (R4) & $0.835$ & $0.843$ & $-0.008$ & -- & -- \\
\bottomrule
\end{tabular}
\end{table}

\paragraph{What the head alone has learned.}
The probe reads the policy $\mathrm{proj}(\text{head}) + w \cdot \text{table}$, and the registered runs use $w = 1$.
Re-reading the forty final checkpoints at $w = 0$ isolates the head. In the visible-rival cell without a channel the
head-only rest
level is $0.178$, $0.186$, $0.305$, $0.287$, $0.489$ under \fixed{} against $0.658$, $0.623$, $0.305$, $0.417$,
$0.596$ under \shuffled{}, a contrast of $-0.231$, negative in four blocks of five and tied at the reported
precision in the third, where the full policy gives
$+0.206$. The head's latent policy runs \emph{against} the persistence effect. This is the behaviour we would expect from a
correction term trained by policy gradient on top of a table that already acts. On the same checkpoints the
head-only swap-opponent effect is $+0.008$ (negative in three blocks of five) against $-0.293$ for the full policy
and $-0.161$ for the zero-head ablation, so the head alone shows no swap drop either, and the D3 residue
is not a partner-dependent policy learned by the language model. On the registered four-cell estimand the head-only contrast is $-0.140$, negative in four blocks of five, against
$+0.274$ for the full policy, so the sign survives averaging over the cells the primary hypothesis averages over.
Exploratory, five blocks, one read-out weight, and this says nothing about what a head trained \emph{under} that
authority would do.

\begin{table}[h]
\centering\footnotesize
\caption{The fifteen predictions A1 to D4, scored twice (C4b and the head read-out are reported with their
paragraphs). D1 and D4 are scored on the three planned blocks of the lowered table weight, and its two added blocks
leave both verdicts unchanged. \emph{Direction} asks whether the point estimate lies where the prediction said. \emph{Interval}
applies the standard of the confirmatory tests. A prediction of no effect holds only
if its 95\% interval lies inside $\pm\mathrm{SESOI}_\Delt = \pm0.10$, and an ordering only if the two intervals do
not overlap. A dash means that no interval exists (single pairs) or that the item is descriptive.}
\label{tab:scorecard}
\setlength{\tabcolsep}{3pt}
\begin{tabular}{@{}l>{\raggedright\arraybackslash}p{4.1cm}>{\raggedright\arraybackslash}p{4.3cm}>{\raggedright\arraybackslash}p{3.7cm}@{}}
\toprule
id & prediction & observed & direction / interval \\
\midrule
A1 & $M = 2$, $\CD \approx 0$ & $-0.032$, $\ci{-0.190}{+0.130}$ & held / inconclusive at 5, precise null at 25 \\
A2 & $\CD(8) \ge \CD(4) > \CD(2)$ & $+0.396$, $+0.363$, $-0.032$ & held / only $\CD(4) > \CD(2)$ \\
A3 & synchronous, $\CD \approx 0$, levels near Nash & $-0.042$, levels $0.29$ and $0.33$ & level clause failed / small negative at 25 \\
A4 & floor lowers $\CD$ & $+0.264$ against $+0.363$ & held / intervals overlap \\
A5x & $10^6$ rounds, two outcomes recorded & \low{} gap persists & not a test \\
C1 & signature $>0$ under \fixed{}, $\approx 0$ under \shuffled{} & Table~\ref{tab:c1} & held / -- \\
C2 & signature grows with $M$ & smaller at $M = 8$ & partly held / -- \\
C3 & slower under \low{} & stronger under \low{} & failed / -- \\
C4 & rematching dose-response & not run as planned & not scorable \\
C5 & $\Enet$ small, unrelated to $\Delt$ & $r = +0.04$, median $|\Enet| = 1.53$ & held / -- \\
C6 & rest points under \shuffled{}/\low{} & listed & descriptive \\
D1 & authority lowers level and contrast & $0.395$ vs $0.673$, $+0.198$ vs $+0.206$ ($+0.235$ on the same blocks) & held / -- \\
D2 & ablation within $0.10$ of reference & $+0.204$ against $+0.206$ & held / -- \\
D3 & ablation swap within $0.10$ of $-0.29$ & $-0.161$ & failed / -- \\
D4 & signature weaker under authority & $+0.028$ against $+0.010$ & failed / -- \\
\bottomrule
\end{tabular}
\end{table}

Scored by direction, seven predictions held, one held in part, four did not and three cannot be scored. Scored by
intervals, none of the predictions of no effect is a precise null at five blocks, and of the orderings only $M = 4$
above $M = 2$ is established. On 25 further blocks (Appendix~\ref{app:tabrev}) the two predictions of no effect,
$M = 2$ and the synchronous learner, hold by the interval rule, the first as a precise null and the second as a small
negative effect whose interval excludes zero. The mechanism evidence is therefore consistent with
spontaneous coupling, and it does not establish it.

\paragraph{Predictions A1 to A5x (committed 14 September 2026, before any run).}
A1, $M = 2$, predicts $\CD \approx 0$, and a substantial $\CD$ would mean the effect comes from the episode clock or the seat
history rather than from the partner's identity. A2, $M = 8$, predicts $\CD(8) \ge \CD(4) > \CD(2)$. A3, synchronous
learner of \citet{asker2022artificial} (every action updated against the observed rival price, with a visible rival
only, since a hidden rival is not observed), predicts $\CD \approx 0$ and levels near Nash in both arms, and a preserved persistence
effect would mean the mechanism is not coupling. A4, permanent exploration floor $\varepsilon \ge 0.05$, predicts $\CD$
smaller than at the zero floor. A5x, $10^6$ rounds under Calvano's schedule, has two recorded outcomes, the
hidden-rival gap persists (the channel is stable at our scale) or closes (the hidden-rival levels at 8\,000 rounds are
transient).

\begin{table}[h]
\centering\small
\caption{Mechanism runs on the tabular learner, per-cell means over five blocks, giving the level averaged over
checkpoints, the level at the final checkpoint $\Delt_{\mathrm{final}}$ (round 8\,000, or round $10^6$ in the last four
rows), $\Enet^{\BRone}$, pattern share and $P_1$.}
\label{tab:mechcells}
\setlength{\tabcolsep}{3.5pt}
\begin{tabular}{@{}llrrrrr@{}}
\toprule
variation & cell & $\Delt$ & $\Delt_{\mathrm{final}}$ & $\Enet^{\BRone}$ & pattern & $P_1$ \\
\midrule
\multirow{4}{*}{$M=4$ (reference)} & \fixed/\high & $0.661$ & $0.773$ & $1.28$ & $0.34$ & $0.17$ \\
 & \shuffled/\high & $0.365$ & $0.308$ & $6.43$ & $0.35$ & $0.07$ \\
 & \fixed/\low & $0.571$ & $0.667$ & $0$ & $0$ & $0$ \\
 & \shuffled/\low & $0.141$ & $0.036$ & $0$ & $0$ & $0$ \\
\midrule
\multirow{4}{*}{$M=2$} & \fixed/\high & $0.688$ & $0.764$ & $1.59$ & $0.45$ & $0.25$ \\
 & \shuffled/\high & $0.721$ & $0.848$ & $3.95$ & $0.36$ & $0.30$ \\
 & \fixed/\low & $0.475$ & $0.513$ & $0$ & $0$ & $0$ \\
 & \shuffled/\low & $0.504$ & $0.614$ & $0$ & $0$ & $0$ \\
\midrule
\multirow{4}{*}{$M=8$} & \fixed/\high & $0.687$ & $0.799$ & $2.39$ & $0.51$ & $0.34$ \\
 & \shuffled/\high & $0.324$ & $0.438$ & $-0.23$ & $0.30$ & $0$ \\
 & \fixed/\low & $0.648$ & $0.721$ & $0$ & $0$ & $0$ \\
 & \shuffled/\low & $0.219$ & $0.105$ & $0$ & $0$ & $0$ \\
\midrule
\multirow{2}{*}{synchronous} & \fixed/\high & $0.291$ & $0.365$ & $1.87$ & $0.14$ & $0$ \\
 & \shuffled/\high & $0.333$ & $0.607$ & $2.76$ & $0.07$ & $0$ \\
\midrule
\multirow{4}{*}{$\varepsilon \ge 0.05$} & \fixed/\high & $0.641$ & $0.645$ & $0.08$ & $0.42$ & $0.20$ \\
 & \shuffled/\high & $0.386$ & $0.492$ & $8.37$ & $0.35$ & $0.06$ \\
 & \fixed/\low & $0.459$ & $0.458$ & $0$ & $0$ & $0$ \\
 & \shuffled/\low & $0.187$ & $0.289$ & $0$ & $0$ & $0$ \\
\midrule
\multirow{4}{*}{$10^6$ rounds} & \fixed/\high & $0.884$ & $0.889$ & $2.71$ & $0.20$ & $0.09$ \\
 & \shuffled/\high & $0.689$ & $0.672$ & $2.82$ & $0.81$ & $0.59$ \\
 & \fixed/\low & $0.685$ & $0.681$ & $0$ & $0$ & $0$ \\
 & \shuffled/\low & $-0.047$ & $-0.090$ & $0$ & $0$ & $0$ \\
\bottomrule
\end{tabular}
\end{table}

Table~\ref{tab:mechcells} gives the per-cell levels behind the five predictions. The reference learner run to
300\,000 rounds gives $\CD = +0.337$, $\ci{+0.265}{+0.425}$. Block contrasts are, for $M = 2$,
$+0.15$, $+0.03$, $-0.19$, $+0.02$, $-0.18$, for $M = 8$, $+0.41$, $+0.31$, $+0.47$, $+0.45$, $+0.34$, for the synchronous
\high{} pair, $+0.01$, $+0.01$, $-0.11$, $-0.08$, $-0.05$, for the floor, $+0.29$, $+0.22$, $+0.29$, $+0.14$, $+0.38$, and for
$10^6$ rounds on the \low{} pair, $+0.77$, $+0.61$, $+0.83$, $+0.75$, $+0.70$. At $10^6$ rounds the two-pair contrast is
$+0.464$, with a confidence set whose hull is $\ci{+0.135}{+0.775}$ (nine grid points inside the hull are rejected), and the \shuffled{}/\high{} cell reaches the highest pattern share of
any tabular cell ($0.81$), so with the rival observed and a million rounds, the rematched learner does eventually
build something the probe reads as punishment, but not the level the persistent one reaches.

\paragraph{Predictions C1 to C6 (committed 14 September 2026, 22:32, before their analyses) and outcomes.}
C1, hybrid agent, blocks 1 to 4, predicts a signature positive under \fixed{}, $\approx 0$ under \shuffled{}, growing over
training, present with a hidden rival as with a visible one, and is \textbf{held} (Table~\ref{tab:c1}). C2, tabular
learner, $M \in \{2,4,8\}$, predicts a signature gap $\approx 0$ at $M = 2$, $> 0$ at $M = 4$, $\ge$ at $M = 8$, and is
\textbf{partly held} ($+0.018$ with a visible rival and $+0.057$ with a hidden one at $M = 4$, $+0.010$ and $+0.026$ at
$M = 8$, where the $M = 8$ gap is smaller, not larger). C3, slower with a hidden rival than with a visible one to a
comparable final level, \textbf{failed}, since the signature is stronger with a hidden rival. C4, dose-response monotone in $p$ with $p = 0 \approx \fixed{}$, is \textbf{not
tested as planned}, see below. C5, $\Enet$ small and uncorrelated with $\Delt$ where coupling is established, is
\textbf{held}. C6 is descriptive.

\begin{table}[h]
\centering\footnotesize
\caption{C1, the coupling signature $\rho(\text{partners}) - \rho(\text{non-partners})$ in the hybrid agent's tables,
blocks 1 to 4, by round, and the two correlations at round 8\,000. The residual variant (correlating each table's
residual around the run's mean table) was defined after seeing these numbers and is post hoc.}
\label{tab:c1}
\setlength{\tabcolsep}{2.5pt}
\begin{tabular}{@{}lrrrrrrrrrrr@{}}
\toprule
cell & 1k & 2k & 3k & 4k & 5k & 6k & 7k & 8k & $\rho_{\text{part}}$ & $\rho_{\text{non}}$ & residual \\
\midrule
\fixed/\high/\none      & $.003$ & $.004$ & $.013$ & $.009$ & $.011$ & $.012$ & $.013$ & $.013$ & $.905$ & $.892$ & $+0.16$ \\
\shuffled/\high/\none   & $-.001$ & $.001$ & $.002$ & $.001$ & $.000$ & $.000$ & $.001$ & $.001$ & $.910$ & $.909$ & $\le 0.12$ \\
\fixed/\high/\freetext    & $.000$ & $.006$ & $.011$ & $.015$ & $.012$ & $.012$ & $.013$ & $.013$ & $.922$ & $.909$ & $+0.24$ \\
\shuffled/\high/\freetext & $.001$ & $.005$ & $.006$ & $.006$ & $.008$ & $.009$ & $.009$ & $.008$ & $.909$ & $.901$ & $\le 0.12$ \\
\fixed/\low/\none       & $.004$ & $.012$ & $.040$ & $.079$ & $.121$ & $.113$ & $.145$ & $.170$ & $.920$ & $.749$ & $+1.04$ \\
\shuffled/\low/\none    & $-.002$ & $.002$ & $.002$ & $-.001$ & $-.002$ & $-.002$ & $.000$ & $-.001$ & $.988$ & $.989$ & $\le 0.12$ \\
\fixed/\low/\freetext     & $.002$ & $.019$ & $.051$ & $.041$ & $.041$ & $.049$ & $.050$ & $.052$ & $.915$ & $.864$ & $+0.54$ \\
\shuffled/\low/\freetext  & $-.002$ & $.000$ & $-.003$ & $-.002$ & $-.003$ & $-.003$ & $-.002$ & $-.003$ & $.982$ & $.985$ & $\le 0.12$ \\
\bottomrule
\end{tabular}
\end{table}

\paragraph{C4 as executed, and C4b.}
The plan said ``$p = 0$: synchronous clock, partner never changed, $\approx$ \fixed{}''. The implementation drew the
initial pairing independently in each of the sixteen markets and froze it at $p = 0$, so each agent kept about
three simultaneous permanent partners. Levels under that variant are, at $p = 0$, $0.39$ / $0.24$ (rival visible /
hidden), at
$0.25$, $0.40$ / $0.20$, at $0.5$, $0.36$ / $0.24$, at $0.75$, $0.45$ / $0.22$, and at $1$, $0.37$ / $0.16$, with the contrast
$p = 0$ minus $p = 1$ at $+0.048$, $\ci{-0.065}{+0.170}$. The error was recorded, the plan rectified in writing before
the relaunch, and C4b run with a fixed initial pairing (A$\leftrightarrow$B on eight markets, C$\leftrightarrow$D on
eight), giving at $p = 0$, $0.64$ / $0.47$, at $0.25$, $0.41$ / $0.21$, at $0.5$, $0.40$ / $0.22$, at $0.75$, $0.41$ / $0.26$, and at
$1$, $0.44$ / $0.21$, with contrast $+0.229$, $\ci{+0.125}{+0.325}$, and a share of ($p_i < p_j$) pairs with
$\Delt_i > \Delt_j$ of $0.71$. The unplanned variant is kept because it is informative, since three permanent partners
give the \shuffled{} level.

\begin{figure}[h]
\centering
\includegraphics[width=0.85\linewidth]{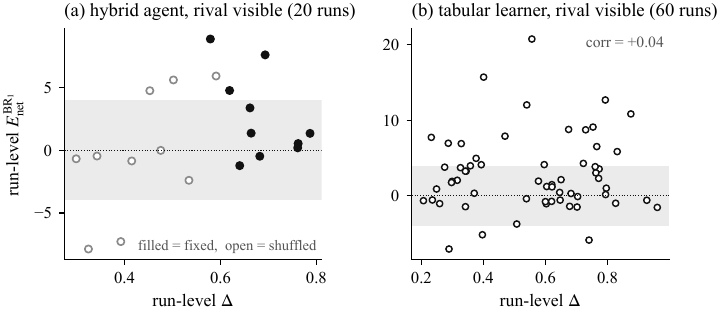}
\caption{Run-level enforcement against run-level level, observable rival. (a) The 20 hybrid runs with a visible rival
(filled for \fixed{}, open for \shuffled{}). (b) The 60 tabular runs of C5. The grey region is $\pm$SESOI$_E$.}
\label{fig:scatter}
\end{figure}

\paragraph{C5 and C6.}
Across the 60 tabular runs with an observable rival (matched baseline, mechanism variants, $10^6$ rounds),
$\mathrm{corr}(\Delt, \Enet^{\BRone}) = +0.04$ and $\mathrm{corr}(\Delt, \Enet^{\BRK}) = +0.03$. Among the 30 runs at
$\Delt > 0.6$ the median $|\Enet^{\BRone}|$ is $1.53$ bins$\cdot$rounds and the median $\Enet$ is $1.22$
(Figure~\ref{fig:scatter}). For C6, the final rest points of \shuffled{}/\low{} at $10^6$ rounds are at grid bins
(6,9), (8,5), (6,5), (4,4), (5,2), (13,4), (3,3), (4,5), (5,4), (5,5) across the five blocks, mostly below the
Nash bin 6, with block levels $+0.06$, $-0.04$, $-0.17$, $-0.19$, $-0.11$, and the \fixed{}/\low{} rests in the same
runs are at bins 11 to 23.

\section{The static-response benchmark, its invariance, dual residuals and validity curve}
\label{app:validity}

\subsection{A reference that sees what the agent sees}
The static reference of Section~\ref{sec:endpoints} best-responds to the rival's exact price, while the agent's state is
the pair of four-wide price buckets. On the $43\%$ of starts whose cut leaves the deviator's own bucket unchanged
(Appendix~\ref{app:explo}), the reference moves and the agent cannot, so $\Enet = \Enf - \Emyo$ is then a mechanical
negative number carrying no behaviour. This is the theorem's case in which the reference is not the filter's internal
function, and it biases $\Enet$ downward. A reference that sees only buckets does not move on those starts. Setting
$\Emyo = 0$ there, and keeping the exact reference elsewhere, with the blind starts identified exactly from the saved
tables of all eight checkpoints, gives block contrasts of $-0.63$, $-0.67$, $+1.23$, $+3.29$ and $+5.36$, a mean
$\CE^{\BRone}$ of $+1.72$ against the registered $+1.48$, positive in three blocks of five as before, and $+3.43$
against $+2.97$ on the visible-rival scale (post hoc). The four hidden-rival cells stay exactly zero. An earlier
version reported a bucket-blind variant built on a crossing flag that compared the cut with the rival's price
instead of the deviator's own, and we withdraw it. A probe with
$k \geq 9$ bins crosses on most starts (84.5\% on the tabular learner, the rest clamped at the bottom of the grid), at
the cost of a different estimand, since the one-shot gain that fixed $k = 4$ no
longer holds. Exploratory, computed from the registered probe files with no run re-executed.

\paragraph{Invariance theorem.}
Let the rival's price in either branch be a causal linear filter of the matched static reference $f$,
$r(t) = \sum_{j \ge 0} h_j\, f(p^{\mathrm{rival}}(t-1-j))$, with $\sum_j |h_j| < \infty$, a common pre-history before
$t^{\ast}$ in both branches, and a summable myopic difference $\mathrm{RE}_{\mathrm{myopic}}$. Then by linearity
$\mathrm{RE}(h) = \sum_j h_j\,\mathrm{RE}_{\mathrm{myopic}}(h-j)$, the sum of a convolution is the product of the
sums, so $\Enf = G\,\Emyo$ with $G = \sum_j h_j$ and
\[
  \Enet = (G-1)\,\Emyo .
\]
Every unit-gain kernel (impulse, pure delay, causal moving average, geometric partial adjustment, and mixtures)
gives $\Enet = 0$ regardless of its shape, and over-reaction is the same statement at $G \neq 1$, and an affine response
$\alpha\,\mathrm{BR} + \beta$ has $G = \alpha$. Verified in the test suite on those kernels and on a sweep of $G$,
with a maximal discrepancy of $4\times10^{-15}$ under the matched reference. At finite horizon the invariance is
asymptotic, and under a non-negative kernel the truncation bias is of demonstrated sign and conservative ($\Enet \le 0$
at $G = 1$, tending to $0$ from below), but a signed kernel can produce a positive truncation residual, so the sign
statement is made only under $h_j \ge 0$.

\paragraph{Why no single reference suffices.}
The fourth condition, that the reference be the filter's own internal function, is the binding one. A rival that
best-responds to the \emph{mean of a window} of past prices filters before the non-linearity, computing
$\mathrm{BR}(\overline{p})$ where the theorem needs $\overline{\mathrm{BR}(p)}$, and leaves a Jensen gap that $\BRK$
absorbs and $\BRone$ does not, while a rival that applies inertia \emph{after} its one-step best response is matched by
$\BRone$ and not by $\BRK$. Measured on this calibration with six unit-gain kernels, the post-$\mathrm{BR}$
inertial family scores $0$ under $\BRone$ and $+1$ bin (one-round deviation) to $+2$ bins (two rounds) under
$\BRK$, an anti-conservative false positive twice the size of the myopic correction itself, while the window family
scores $-1$ to $-2$ bins under $\BRone$ and $0$ under $\BRK$. The two leftover biases mirror each other and have
opposite signs, so H6 rejects only if both references reject, at unchanged $\alpha$. A sweep over
effective window $M \in \{1,\dots,12\}$, deviation durations $\{1,2,3,4,6\}$, with and without an added unit-gain
post-$\mathrm{BR}$ kernel (192 combinations) produced no case in which both components were positive, while a
simulated grim trigger passes both with more than 100 bins$\cdot$rounds. The residual is not bounded by a constant, since
under the unmatched reference it is about $-\min(M-1, L)$ bins for a deviation persisting $L$ rounds, measured to
$-6$, and in the contrast $\CE$ its sign follows the difference in realised persistence between arms, which is why
the deviator's persistence $L$ is reported per cell (Table~\ref{tab:percell}) and why $\mathrm{SESOI}_E$ was set
above the resulting band.

\paragraph{Validity curve.}
Whether the correction matters is a property of the demand system, and it can be computed with no rollout. With
$d^{\ast}(\mu, m_k)$ the smallest undercut of the monopoly price, as a fraction of the Nash-to-monopoly gap, that
yields a one-shot gain of at least $m_k = 10\%$ (the rule that fixed $k = 4$ here), define
\[
  R(\mu) = \frac{\left|\mathrm{BR}(p_{\mathrm{mon}}) - \mathrm{BR}(p_{\mathrm{mon}} - d^{\ast}\,(p_{\mathrm{mon}}
  - p_{\mathrm{Nash}}))\right|}{p_{\mathrm{mon}} - p_{\mathrm{Nash}}},
\]
the static response to a calibrated deviation as a share of the amplitude a grim trigger would move the rival by.
Table~\ref{tab:validity} gives the curve. Two facts were not expected. $R$ \emph{increases} with differentiation
although the slope of $\mathrm{BR}$ decreases, because a profitable deviation must be deeper when products are more
differentiated and $R \approx d^{\ast} \cdot \partial\mathrm{BR}/\partial p$ is dominated by the depth. And the
deviation rule has no solution beyond $\mu^{\ast} = 0.50046$, since past that differentiation, no undercut of the
monopoly price gains 10\% in one shot, so a forced-deviation test loses its meaning. Grid quantisation does not
inflate the correction, since at $\mu = 0.25$ the discrete value at 20 Nash-to-monopoly steps is $R = 1/20 = 0.050$
exactly (one bin), against $0.0625$ at 80 and 160 steps and $0.0622$ in the continuum. $R$ is a ratio of
instantaneous amplitudes around the monopoly price, and it is neither a ratio of areas nor a bound on $\Emyo/\Enf$ on
the states a trained policy actually visits, which is what Table~\ref{tab:percell} measures.

\begin{table}[h]
\centering\small
\caption{The validity curve, continuous (no grid), $m_k = 0.10$. The slope is $\partial\mathrm{BR}/\partial p_{\mathrm{rival}}$
at the monopoly price, and $d^{\ast}$ the required deviation depth as a fraction of the Nash-to-monopoly gap. The
calibration used in this paper is $\mu = 0.25$, where the discrete grid gives $R = 0.050$.}
\label{tab:validity}
\begin{tabular}{@{}lrrrrrrrrr@{}}
\toprule
$\mu$ & $0.05$ & $0.10$ & $0.20$ & $0.25$ & $0.30$ & $0.40$ & $0.45$ & $0.49$ & $0.50$ \\
\midrule
slope & $0.835$ & $0.674$ & $0.436$ & $0.357$ & $0.299$ & $0.222$ & $0.196$ & $0.179$ & $0.175$ \\
$d^{\ast}$ & $0.013$ & $0.035$ & $0.110$ & $0.166$ & $0.234$ & $0.409$ & $0.530$ & $0.681$ & $0.774$ \\
$R$ & $0.011$ & $0.024$ & $0.050$ & $0.062$ & $0.074$ & $0.098$ & $0.114$ & $0.136$ & $0.153$ \\
\bottomrule
\end{tabular}
\end{table}

\paragraph{Predictions D1 to D4 (committed 15 September 2026, before any run) and outcomes.}
Two experiments on the trained agent, in the visible-rival cell without a channel, arms \fixed{} and \shuffled{}, with the block and
slot seeds of the confirmatory runs, 8\,000 rounds, the same probes. D1, authority (table weight $5.5$ instead of
$16.5$), predicts a smaller contrast than $+0.206$ and a \fixed{} level below $0.623$, and is \textbf{held}, though only its
level clause is strong, the contrast moving from $+0.235$ to $+0.198$ on the same three blocks, which is inside the
noise. D2, matched ablation (zero frozen head, frozen backbone), predicts contrast and arm levels within $0.10$ of the
reference, and is \textbf{held} ($+0.204$, with $0.681$ and $0.477$, and per-block paired differences averaging $+0.009$ and
$+0.010$, reaching $\pm0.18$). D3, ablation swap effect within $0.10$ of $-0.29$, \textbf{failed} ($-0.161$, with
blocks $-0.25$, $-0.22$, $-0.03$, $+0.03$, $-0.34$). D4, coupling signature positive under \fixed{} and near zero
under \shuffled{} in the ablation, and weaker under authority, \textbf{failed}, since the ablation part holds
($+0.010$ and $-0.003$) but the authority variant's signature is larger, not smaller ($+0.028$ on the three planned
blocks, and $+0.117$ on the two added ones).

The authority experiment was limited to blocks 1 to 3 for time, written down before any of its results existed, and
D1 and D4 were scored on those three blocks, whose contrasts are $+0.162$, $+0.143$ and $+0.288$. Blocks 4 and 5 were
added on 23 September after reading them, with two predictions committed before their runs, F1, a positive contrast
in both new blocks, and F2, a five-block mean of at least $+0.10$. Both held (block contrasts $+0.229$ and $+0.133$,
and $+0.191$ for the five-block mean that F2 names). The runs finished on 24 September and were read then. We report
these two blocks apart from the planned ones (Table~\ref{tab:parts}), and they leave the verdicts of D1 and D4
unchanged. The head overturns the table in 49\% of settled seats on the planned blocks, and in 298 of 600 over all
checkpoints of the ten runs.

A table weight of $3$ was planned on 23 September at 22:01, after blocks 1 to 3 at $5.5$ and the first checkpoint of
the language model alone (a training contrast of $+0.01$ at round 1\,000) had been read, and amended at 22:54,
before any of its runs, to five blocks, with G1 dropped (the override rate became a manipulation check) and G2 and
G3 replaced by three predictions on levels averaged over checkpoints, G2$'$, a contrast at least $0.05$ below the
one at $5.5$ on the same blocks, G3$'$, a positive slope of the block contrasts on the table's weight across
$16.5$, $5.5$ and $3$, and G4, a \fixed{} arm no higher than its starting level of $0.39$, read as rematching
lowering prices rather than persistence raising them. The original G2, a mean contrast below $+0.198$ on blocks 1
to 3, would also have failed ($+0.228$). On 24 September at 10:14 we set a read time, 25 September at 12:00, that
would keep only blocks whose two arms had finished. All ten runs finished their probes at 09:06 and were read once
at 09:08, before that time, the same set of blocks the read time would have kept. The block contrasts are $+0.272$, $+0.132$, $+0.280$, $+0.076$ and $+0.035$, mean $+0.159$,
against $+0.191$ at $5.5$ and $+0.206$ at $16.5$ on the same blocks, with arms at $0.351$ and $0.192$. G2$'$ failed
(a gap of $0.032$), G3$'$ held (a slope of $+0.0028$ per unit of weight) and G4 held. The head overturns the table in
695 of 884 settled seats (79\%) over all checkpoints of the ten runs, against 49\% at $5.5$.

The
ablation runs used a code path that skips the language model's forward passes, which cannot change an action when
the head is zero and frozen, and it was validated by requiring every logit vector of a short run to equal the full
path's, and by a probe whose output on the same checkpoint is identical.

\section{Implementation notes on the inference}
\label{app:inference}
These are properties of the registered procedure that the registration did not spell out, and the statistic, basis and
grids are unchanged.
\begin{itemize}[leftmargin=1.4em,itemsep=1pt]
\item The block-level statistic is \textbf{studentised}, that is scaled by its own standard error. An unscaled test
  is exact only under the sharp null that no unit's outcome depends on persistence. Block effects and the effects of
  observability and channel cancel inside each (\fixed{}, \shuffled{}) difference and do not break that null, but a
  persistence effect that varies across seeds or cells does, as our post hoc breakdown by observability suggests.
  Studentising keeps exactness under the sharp null and is meant to add validity for the weak null of a zero average
  effect as the number of blocks grows. \citet{chung2013exact} prove this for two-sample and $k$-sample permutation
  tests, not for the sign-flip test used here, and at five blocks it is approximate in any case.
\item The inverted confidence set is \textbf{not guaranteed to be contiguous} at these block counts, because the
  $p$-value surface over $\tau$ is coarse. We report the hull of the retained $\tau$, the number of grid points
  inside the hull that the test rejects (zero for every interval in this paper except the $10^6$-round two-pair
contrast of Appendix~\ref{app:mechanism}, where nine are), and a flag raised when the set reaches the edge of the grid. The
  compatible sharp null is completed as $z_c = (\tau/2)\,P_c$ with all six other contrasts at zero, and ``every unit
  has effect $\tau$'' does not by itself pin down the potential outcomes of an eight-cell factorial.
\item For the one-sample sign-flip basis used in stage 2 and in the exploratory contrasts, studentisation
  \textbf{cannot change} a $p$-value, because $\sum_i (v_i - \tau)^2$ is invariant under sign flips, so
  $T^2 = n(n-1)m^2/(S - nm^2)$ is strictly increasing in $m^2$ and the ordering of the $2^n$ patterns is that of
  the raw means. Those $p$-values are computed from the means, which is also the numerically stable route since
  $S - nm^2$ is a catastrophic cancellation whenever $|m| \gg \mathrm{sd}$. At the block level the invariance fails
  and studentisation does change the ordering, which is where it is registered.
\item Ties in the reference distribution (a pattern and its global sign flip have equal $|T|$) are counted with a
  relative tolerance, and without one the $p$-value is anti-conservative by $2^{-n}$.
\item A single (\fixed,\shuffled) pair at five blocks admits $2^5$ sign patterns and a smallest two-sided $p$ of
  $0.0625$, so no 95\% interval exists for single-pair contrasts, and they are reported as point estimates with their
  block values.
\end{itemize}

\paragraph{Power of H6.} The registration simulated the test with the pilot's run-to-run spread of $\Enet^{\BRone}$
and gave a power of $0.53$ at $\mathrm{SESOI}_E$, which bounds the joint test of both benchmarks. After the analysis
we recomputed it by re-centring the twenty observed pair differences, flipping their signs at random and adding a
true effect, which gives $0.48$ and $0.47$ for the two benchmarks (post hoc).